\pdfoutput=1
\documentclass[10pt,twocolumn,letterpaper]{article}

\usepackage{cvpr}              

\definecolor{cvprblue}{rgb}{0.21,0.49,0.74}
\usepackage[accsupp]{axessibility}
\usepackage[pagebackref,breaklinks,colorlinks,allcolors=cvprblue]{hyperref}
\usepackage{booktabs}  
\usepackage{tabularx}
\usepackage{multirow}  
\usepackage{graphicx}  
\usepackage{array}     
\def\paperID{3860} 
\def\confName{CVPR}
\def\confYear{2025}
\usepackage{xcolor}  
\usepackage{adjustbox}
\usepackage{float}
\usepackage{placeins}
\usepackage{pgfplots}
\usepackage{pgfplotstable}
\usepackage{capt-of}
\usepackage{flushend, cuted}
\newcommand{\rf}[1]{\textcolor{red}{#1}}   
\newcommand{\bd}[1]{\textcolor{blue}{#1}}  
\title{ Dual Prompting Image Restoration with Diffusion Transformers}
\newcommand\blfootnote[1]{%
  \begingroup
\renewcommand\thefootnote{}\footnote{#1}%
  \addtocounter{footnote}{-1}%
  \endgroup
}

\author{Dehong Kong$^{1,2,*}$, Fan Li$^{3,*,\dag}$,Zhixin Wang$^{3}$,
Jiaqi Xu$^{4}$, Renjing Pei$^{3}$, Wenbo Li$^{3}$, WenQi Ren$^{1,2,5,\dag}$\\
\small{$^1$ School of Cyber Science and Technology, Shenzhen
Campus of Sun Yat-sen University}\\
\small{$^2$MoE Key Laboratory of Information Technology} \quad
\small{$^3$Huawei Noah’s Ark Lab}\quad
\small{$^4$	The Chinese University of Hong Kong}\\
\small{$^5$Guangdong Provincial Key Laboratory of Information Security Technology}\quad
}

\begin{document}
\maketitle

\begin{strip}

\centering
\vspace{-40pt}
\includegraphics[width=\textwidth]{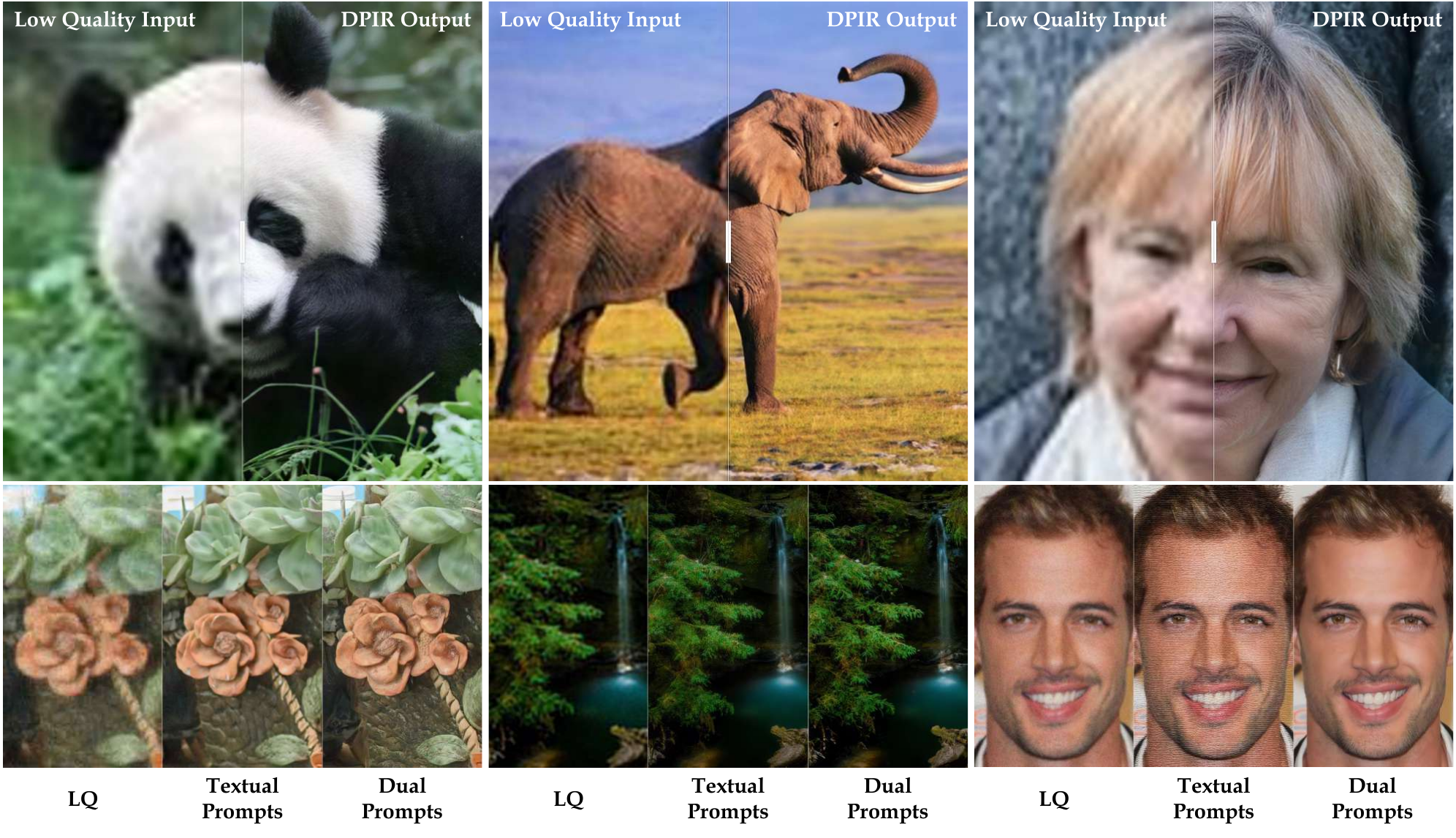}
\vspace{-7mm}

\captionof{figure}{
DPIR exhibits excellent restoration performance on real low-quality images. We compare low quality input, and the results of models using only textual prompts, and our visual-text dual prompts in the lower part. The proposed dual prompting strategy consistently outperforms the single text prompting variant in terms of image restoration quality and fedility.
}
\label{fig:teaser}
\end{strip}

\begin{abstract}
\vspace{-18pt}
\blfootnote{
\noindent $^*$ Contribute Equally. \quad $^\dag$ Corresponding Author.
}

Recent state-of-the-art image restoration methods mostly adopt latent diffusion models with U-Net backbones, yet still facing challenges in achieving high-quality restoration due to their limited capabilities. Diffusion transformers (DiTs), like SD3, are emerging as a promising alternative because of their better quality with scalability. In this paper, we introduce DPIR (Dual Prompting Image Restoration), a novel image restoration method that effectivly extracts conditional information of low-quality images from multiple perspectives. Specifically, DPIR consits of two branches: a low-quality image conditioning branch and a dual prompting control branch. The first branch utilizes a lightweight module to incorporate image priors into the DiT with high efficiency. More importantly, we believe that in image restoration, textual description alone cannot fully capture its rich visual characteristics. Therefore, a dual prompting module is designed to provide DiT with additional visual cues, capturing both global context and local appearance.
The extracted global-local visual prompts as extra conditional control, alongside textual prompts to form dual prompts, greatly enhance the quality of the restoration. Extensive experimental results demonstrate that DPIR delivers superior image restoration performance.
\end{abstract}    
\section{Introduction}
\label{sec:intro}
The persistent demand for high-quality images across diverse fields, such as digital art and medical imaging, drives research in image restoration and super-resolution.
Image restoration (IR) aims to reconstruct high-quality (HQ) images from low-quality (LQ) inputs, addressing real-world degradations of varying complexity, such as noise, blur, and compression artifacts.
Recently, diffusion models-based methods~\cite{wang2024sinsr,he2024one,wu2024one,wang2024exploiting,xie2024addsr,wu2024seesr} have shown the advantages of leveraging the powerful pre-trained text-to-image (T2I) models to enhance the performance of real-world IR. These existing IR methods rely on latent diffusion models~\cite{rombach2022high} with the U-Net~\cite{ronneberger2015u} architecture, which, while effective, still encounter challenges in restoration quality. 
As an alternative, diffusion transformers~\cite{DiTs} (DiTs) recently show promising generative capabilities due to their ability to capture long-range dependencies and their scalability potential, which are crucial for the quality and fidelty of restoration across a wide range of real-world situations.

Despite these advantages, effectively incorporating LQ information into DiTs remains underexplored.
Existing IR methods~\cite{lin2023diffbir,yu2024scaling} based on the U-Net backbone typically utilize ControlNet~\cite{zhang2023adding}, which incorporates the LQ image by employing a trainable copy of the U-Net's encoding layers.
In parallel, StableSR~\cite{wang2024exploiting} adopts a lightweight trainable control network, akin to the T2I Adaptor~\cite{mou2024t2i}, but suffers from moderate performance.
However, these conditional control methods are not optimized for DiTs composed of ViT~\cite{ViTs} blocks.
This discrepancy necessitates a careful design of conditioning mechanisms for DiTs in the IR task.

In this work, we present a dual prompting image restoration model (DPIR), combined with a DiT-based diffusion model Stable Diffusion 3 (SD3)~\cite{esser2024scaling}.
We propose several key techniques in DPIR to effectively integrate control signals from LQ images into the DiT, enabling natural and faithful image restoration.
DPIR includes a low-quality image conditioning branch and a dual prompting control branch, which balances model efficiency and effectiveness.
Specifically, the first branch is a lightweight module designed to efficiently inject the LQ image prior into the DiT. Inspired by ControlNeXt~\cite{peng2024controlnext}, this branch extracts conditional features of LQ images by a few of convolution layers rather than heavy backbone copies like ControlNet. Notably, the lightweight module and the DiT backbone are jointly trained to handle diverse real-world scenarios.

In IR tasks, visual cues of LQ images, including the global context information and the local texture infomation, are crucial for desirable image restoration. 
These visual signals cannot be fully captured by text descriptions alone, especially considering that the DiT backbone, without a skip-connection mechanism like U-Net, hardly keeps the information of input LQ images, which are the conditioning in text-to-image models across all DiT blocks.
Hence, we design a dual prompting restoration branch to incorporate LQ information into DiT from a more comprehensive visual perspective, leveraging the knowledge of pre-trained T2I models and complementing our lightweight conditioning branch.
In detail, given a cropped patch of the LQ image, we first feed it into CLIP ~\cite{radford2021learning} to extract local visual features.
Additionally, global visual features from the surrounding regions, rich in contextual information, are also extracted.
These global-local features function as visual tokens, replacing the original CLIP text features in SD3 and serving as conditioning, alongside T5~\cite{raffel2020exploring} text prompts as visual-text dual prompts, for each block of DiT.

As shown in Figure~\ref{fig:teaser}, DPIR with the dual prompting strategy demonstrates excellent restoration performance and shows a clear advantage over the simple text prompts as DiT conditioning.
Additionally, to further exploit the scalability of DiTs, we use over 20 million high-quality images as training data. Experimental results validate the effectiveness of the proposed DPIR, showing compelling image restoration performance compared to recent state-of-the-art methods and achieving impressive results across a broad range of image restoration tasks.



\begin{figure*}[ht]
\setlength{\abovecaptionskip}{0.1cm}
\setlength{\belowcaptionskip}{-0.2cm}
	\scriptsize
	\centering
    \includegraphics[width=\textwidth]{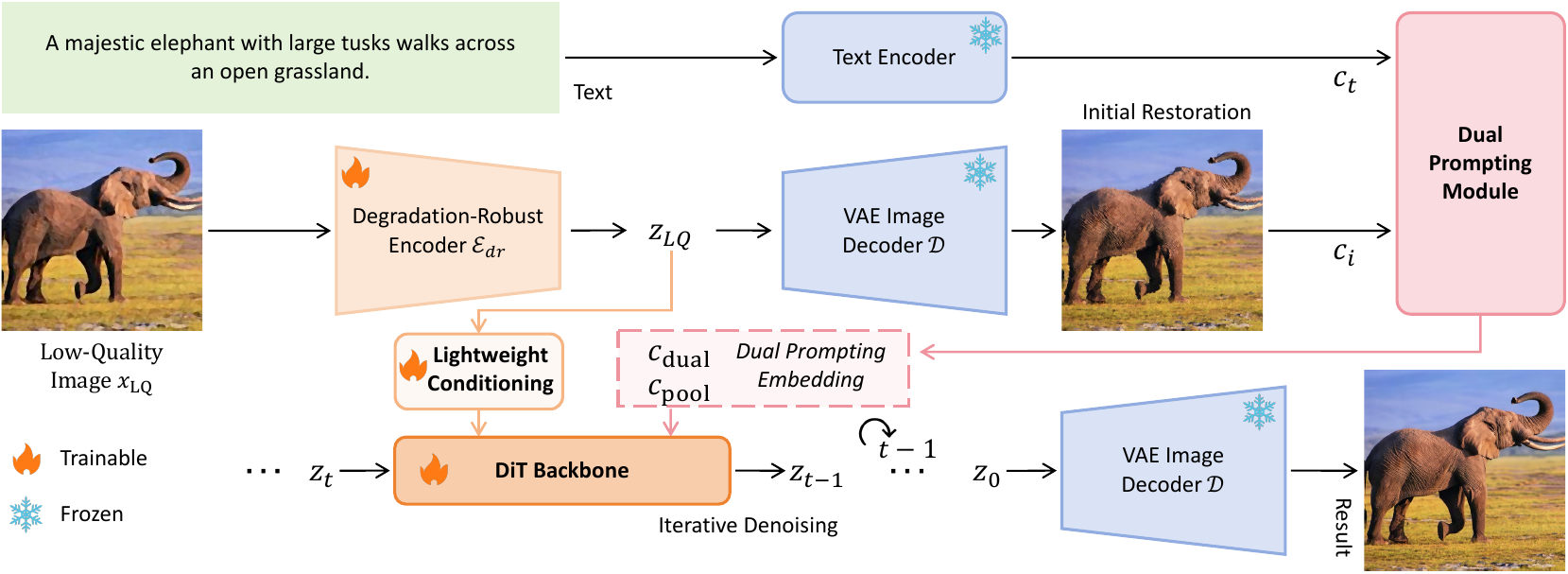}
	\vspace{-2mm}
	\caption{
    Framework of our proposed DPIR.
    Given a low-quality (LQ) image, a lightweight conditioning branch efficiently introduces the LQ information into the DiT backbone. Additionally, a dual prompting restoration branch extracts global and local visual information, alongside text prompts, to form visual-text dual prompts, which greatly enhances the restoration quality and fidelity.}
    \label{fig:pipeline}
    \vspace{-2mm}
\end{figure*}

\section{Related Work}

\paragraph{Image Restoration.}
IR aims to recover high-quality, degradation-free images from degraded inputs~\cite{fan2020neural,jinjin2020pipal,zhang2022accurate}.
Early research focus on addressing specific types of image degradation, such as denoising, deblurring, and super-resolution (SR).
Since the pioneering work of SRCNN~\cite{dong2014learning}, numerous deep learning-based approaches~\cite{chen2023activating,liang2021swinir,shi2016real,zhang2017beyond,johnson2016perceptual,yue2024deep,zhang2018image,lim2017enhanced,zhang2018unreasonable}, have been developed.
Recently, generative model-based IR methods are evolving, including GAN-based and diffusion-based methods, to tackle real-world IR tasks involving blind and complex degradations. DACLIP~\cite{luo2023controlling} transfers pretrained vision-language models to low-level vision tasks as a multi-task framework for image restoration.

\vspace{-4mm}
\paragraph{GAN-based IR.}

The application of GANs for IR or SR dates back to SRGAN~\cite{ledig2017photo}, where image degradation is modeled using bicubic downsampling.
Subsequently, BSRGAN~\cite{zhang2021designing} and Real-ESRGAN~\cite{wang2021real} show that IR GANs can handle more complex image degradations by synthesizing more realistic HQ-LQ pairs, showing convincing results in real-world IR.

\vspace{-4mm}
\paragraph{Diffusion-based IR.}
Early diffusion-based works~\cite{kawar2021snips,kawar2022denoising,wang2022zero} use pre-trained diffusion models~\cite{ho2020denoising,song2020score,dhariwal2021diffusion} to tackle the IR problem with simple degradation modeling.
Recent approaches \cite{wang2024exploiting,wu2024seesr,yang2023pixel,lin2023diffbir,yu2024scaling,lu20243d} use advanced pre-trained T2I models, such as Stable Diffusion~\cite{rombach2022high}, for real-world image restoration.
These methods typically introduce adapters to incorporate LQ images as control signals into diffusion models to reconstruct HQ images.
StableSR~\cite{wang2024exploiting} redesigns a time-aware encoder with LQ embedding.
SUPIR~\cite{yu2024scaling} capitalizes on the generative capability of SDXL~\cite{podell2023sdxl} and the captioning strength of LLaVA~\cite{liu2024visual} to synthesize rich image details.
In parallel, other works~\cite{yue2023resshift,wang2024sinsr} formulate customized restoration diffusion models to tackle IR, but they still exhibit suboptimal performance.

\vspace{-4mm}
\paragraph{Conditional Control.}
Adding spatial conditional control allows powerful T2I models to generate images that follow the appearance or structures of the input reference images.
ControlNet~\cite{zhang2023adding} proposes to use trainable copies and zero convolution to introduce conditioning.
T2I-Adapter~\cite{mou2024t2i} instead leverages a simple and lightweight convolutional network for the controlling.
ControlNeXt~\cite{peng2024controlnext} designs an efficient control architecture and cross-normalization to facilitate training efficiency.
These works target general conditional signals, such as edge, pose, or depth. 
In this work, we focus on designing effective conditional control methods for image restoration requiring high quality and fidelity.

\section{Preliminaries}
\label{Sec:Preliminaries}

\noindent \textbf{Diffusion and Flow.}
Diffusion models~\cite{DDPM, song2021score, DDIM} assume a forward noising process following a specific Markov chain.
We relax this and consider general generative modeling that defines a mapping~\cite{esser2024scaling} between samples from a noise distribution $\pi_1$ to samples $x_0$ from a data distribution $\pi_0$.
The forward process $p_t$ between two distributions can be expressed as $z_t = a_tx_0 + b_t\epsilon$,
where $\epsilon \sim \mathcal{N}(\mathbf{0}, \mathbf{I})$.


Recent studies, including Stable Diffusion 3 (SD3)~\cite{esser2024scaling}, show that flow matching~\cite{lipman2022flow} or rectified flow~\cite{liu2022flow} formulation exhibits superior performance compared to previously established diffusion formulations, particularly in high-resolution image synthesis.
We follow the same rectified flow formuation, which defines a straight path between the noise distribution and the data distribution: $z_t = (1-t)x_0 + t\epsilon$.
During training, we optimize the network $v_\theta$ with the conditional flow matching objective:
\begin{equation}
    \mathcal{L}_{CFM} = \mathbb{E}_{t,p_t(z|\epsilon),p(\epsilon)} \left[\vert\vert v_\theta(z_t,t) - (\epsilon - x_0) \vert\vert^2 \right]
\end{equation}
Note that $v_\theta$ can take additional inputs besides $z_t,t$, such as text or other control signals.

\vspace{-4mm}
\paragraph{Diffusion Transformers (DiTs).}
Earlier, latent diffusion models (LDMs)~\cite{rombach2022high,podell2023sdxl} for image generation adopt the U-Net backbone, while diffusion transformers~\cite{DiTs}, which replace the U-Net backbone~\cite{ronneberger2015u} with vision transformers (ViTs)~\cite{ViTs}, are becoming an increasing trend.
DiTs show that ViT can serve as scalable architectures for diffusion models with respect to the network complexity vs. sample quality.
SD3\footnote{https://huggingface.co/stabilityai/stable-diffusion-3-medium}~\cite{esser2024scaling} is a Multimodal Diffusion Transformer (MMDiT) text-to-image model that features greatly improved performance in image quality and resource efficiency, achieved through extensive training on high-quality aesthetic images.
In this work, we propose to explore the power of SD3 to further advance image restoration.

\section{Method}

In this work, we propose DPIR, a novel image restoration model based on SD3, which consists of a low-quality image conditioning branch and a dual prompting restoration branch. Figure~\ref{fig:pipeline} shows the pipeline of DPIR.

\subsection{Overview of DPIR}

As shown in Figure~\ref{fig:pipeline}, the input low-quality image $x_{\text{LQ}}$ is first fed to the degradation-robust VAE encoder $\mathcal{E}_{dr}$ to obtain the LQ image condition $z_{\text{LQ}}$ (Sec.~\ref{subsec:drvae}).
Then, $z_{\text{LQ}}$ is incorporated into the DiT via a lightweight low-quality image conditioning branch (Sec.~\ref{subsec:lqicb}).
In parallel, $z_{\text{LQ}}$ is mapped back to the pixel space using the VAE decoder $\mathcal{D}$, which together with the given textual prompts, are sent to another branch, \ie, the dual prompting module, to obtain the dual prompting embeddings (Sec.~\ref{subsec:dpcb}).
The conditional signals from these two branches guide the DiT to generate $z_0$, the latents of HQ images, through iterative denoising.
Moreover, we propose a global-local visual prompting training strategy to improve the restoration learning (Sec.~\ref{subsec:dpcb}).
The learnable parameters, including the lightweight module of the LQ image conditioning branch, MLP layers of the dual prompting branch, and all DiT blocks, are jointly trained using the conditional flow matching objective, which is capable to exploit the scalability of DiTs.

\subsection{Degredation-Robust VAE Encoder}
\label{subsec:drvae}
We fine-tune the VAE encoder of SD3 to make our latent LQ condition, $z_{\text{LQ}}$, more robust to degradation, inspired by SUPIR~\cite{yu2024scaling}.
Unlike SUPIR, we add LPIPS and GAN losses to prevent the VAE from producing overly smooth results, thereby preserving details for input images with relatively high quality.
Note that the SD3 VAE with 16 latent channels outperforms SDXL VAE~\cite{podell2023sdxl} with 4 channels, providing a better initial LQ condition for our DiT.
The fine-tuning of $\mathcal{E}_{dr}$ optimizes the following loss:
\begin{equation}
    \vert\vert\mathcal{D}(\mathcal{E}_{dr}(x_{\text{LQ}})) - x_{\text{HQ}})\vert\vert_1 + \alpha \mathcal{L}_{lpips} + \beta \mathcal{L}_{GAN} \ ,
\end{equation}
where $\alpha$ and $\beta$ are weighting hyperparameters, and $x_{\text{GT}}$ is the ground truth HQ image.

\begin{figure}[tp]
    \centering
	\scriptsize
    \includegraphics[width=0.9\hsize]{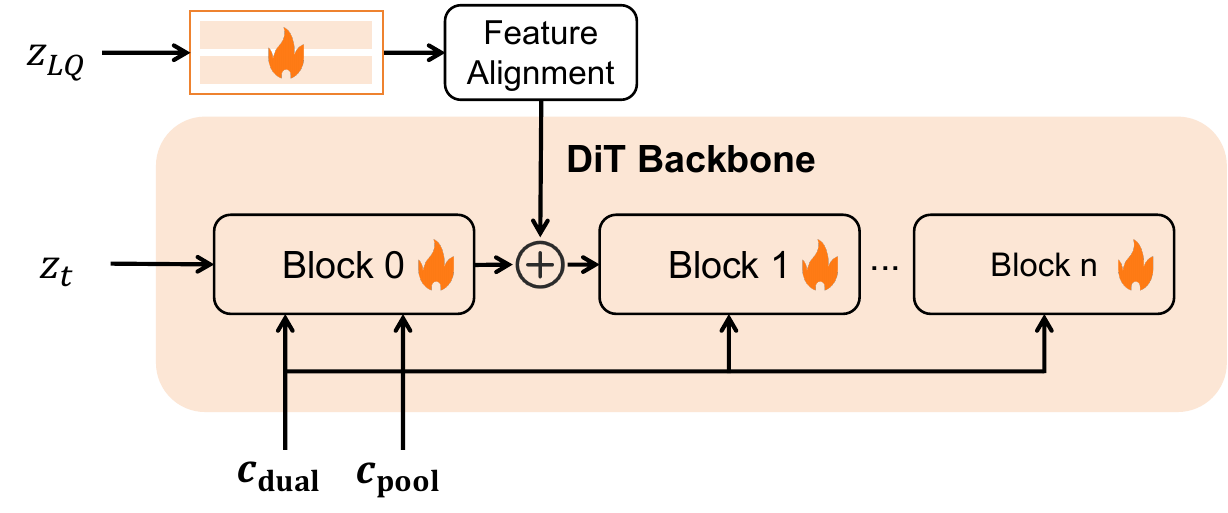}
	\vspace{-4mm}
	\caption{
    The LQ conditioning branch has a lightweight feature extraction module and an adaptive feature alignment module.
    }
    \label{fig:controlnext}
	\vspace{-4mm}
\end{figure}

\subsection{Low-Quality Image Conditioning Branch}
\label{subsec:lqicb}
The low-quality image conditioning branch is a lightweight module that integrates LQ conditioning $z_{\text{LQ}}$ into the DiT backbone.
The structure of this efficient image conditioning branch of DPIR is illustrated in Figure~\ref{fig:controlnext}.
Inspired by the emerging controllable generation approach, ControlNeXt~\cite{peng2024controlnext}, the LQ conditioning branch consists of two main components: a lightweight feature extraction module with a few trainable convolutional layers $\mathcal{F}_c(\cdot ;\phi_c)$, and an adaptive feature alignment module $\eta$.
The conditional control is incoporated into DiT with parameters $\theta_d$ by encoding the LQ features, normalizing the encoded features, and adding them to the output of DiT's first block with parameters $\theta_{d_0}$ as follow:
\begin{equation}
    y_c = \mathcal{F}_{d_0}(z_t; \theta_{d_0})
    + \eta(\mathcal{F}_c(z_{\text{LQ}}; \phi_c); \mu, \sigma ) \ ,
\end{equation}
where $\theta_{d_0} \subseteq \theta_d, \phi_c \ll \theta_d$, $z_t$ denotes the noisy latent at timestep $t$, and $\mu$ and $\sigma$ are the measured mean and variance of the output features from DiT's first block $\mathcal{F}_{d_0}(z_t; \theta_{d_0})$.

The adaptive feature alignment function $\eta$ aligns the conditional features with the main branch features of DiT in terms of mean and variance, normalizing them to ensure stable and effective training.

\begin{figure}[tp]
    \centering
    \includegraphics[width=0.9\hsize]{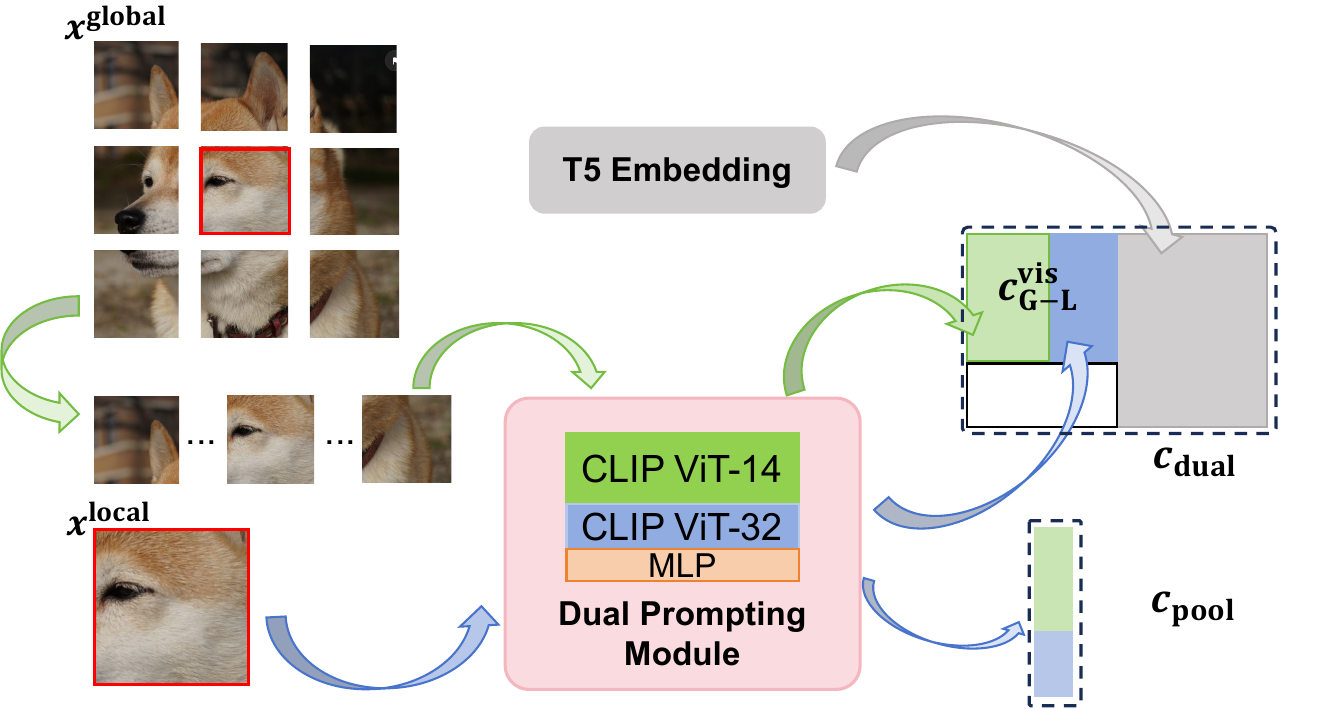}
    \caption{This figure shows a dual prompting control branch and the generation of dual embedding and cls embedding.}
    \vspace{-4mm}
    \label{fig:dual}
\end{figure}

	
    

\begin{figure*}[t]
    \centering
    \begin{subfigure}[b]{0.16\hsize}
        \centering
        \includegraphics[width=\hsize]{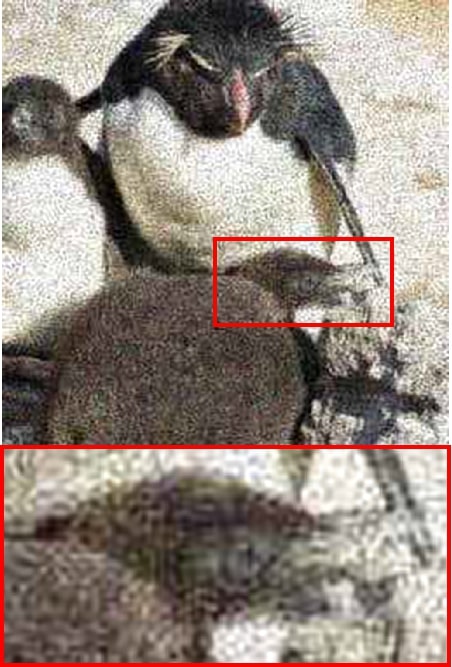}
   
    \end{subfigure}
    \begin{subfigure}[b]{0.16\hsize}
        \centering
        \includegraphics[width=\hsize]{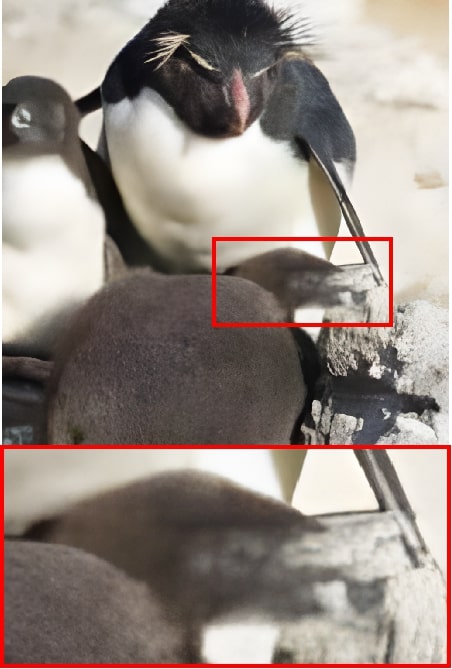}
      
    \end{subfigure}
    \begin{subfigure}[b]{0.16\hsize}
        \centering
        \includegraphics[width=\hsize]{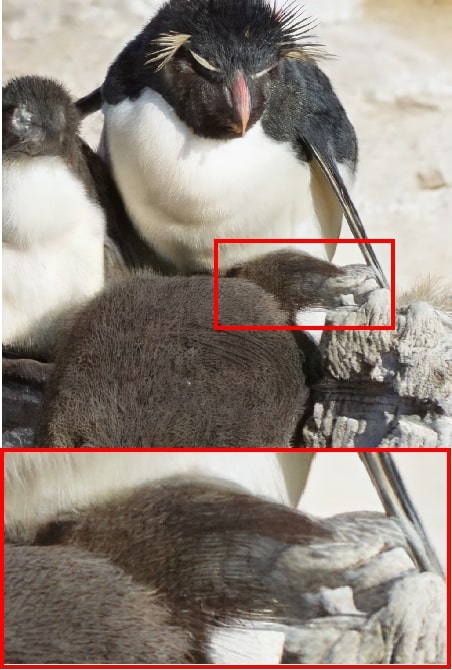}
 
    \end{subfigure}
    \begin{subfigure}[b]{0.16\hsize}
        \centering
        \includegraphics[width=\hsize]{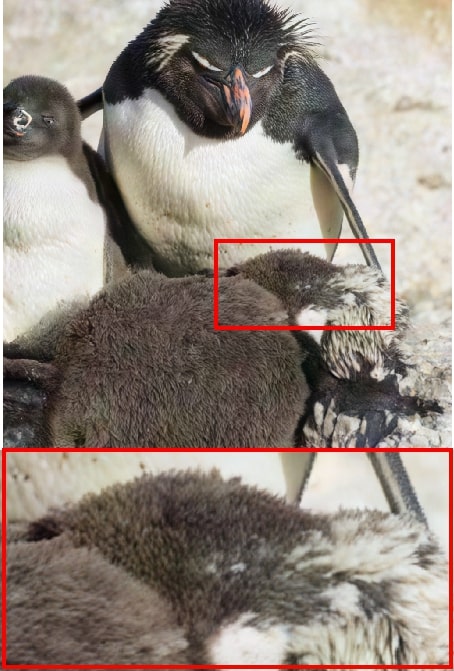}
 
    \end{subfigure}
    \begin{subfigure}[b]{0.16\hsize}
        \centering
        \includegraphics[width=\hsize]{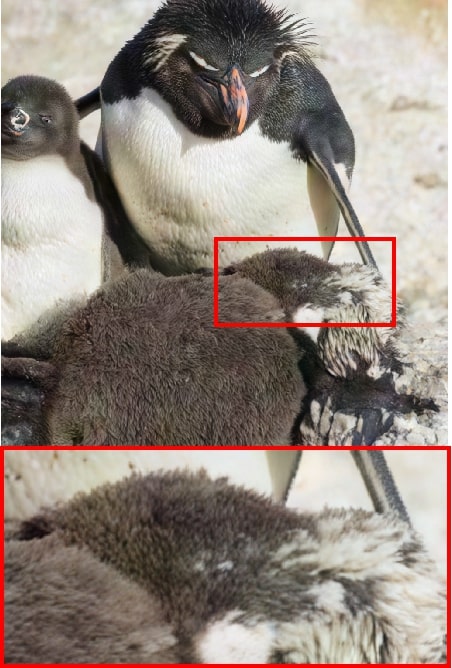}
 
    \end{subfigure}
    \begin{subfigure}[b]{0.16\hsize}
        \centering
        \includegraphics[width=\hsize]{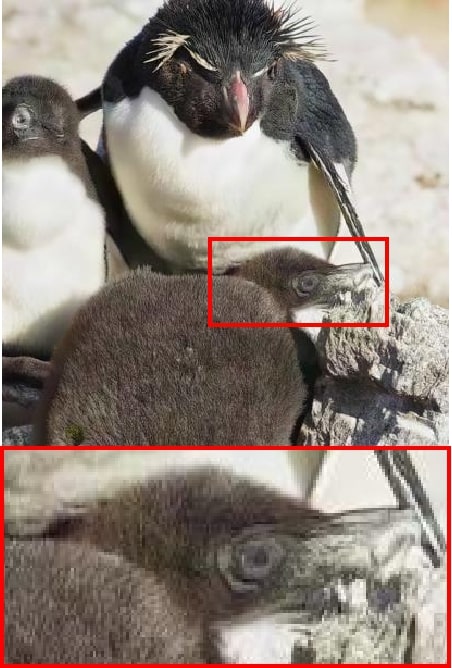}
 
    \end{subfigure}
    
    \vspace{-0.3cm} 
    \begin{subfigure}[b]{0.16\hsize}
        \centering
        \includegraphics[width=\hsize]{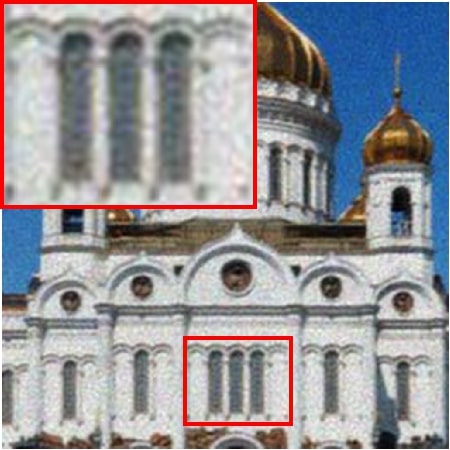}
   
    \end{subfigure}
    \begin{subfigure}[b]{0.16\hsize}
        \centering
        \includegraphics[width=\hsize]{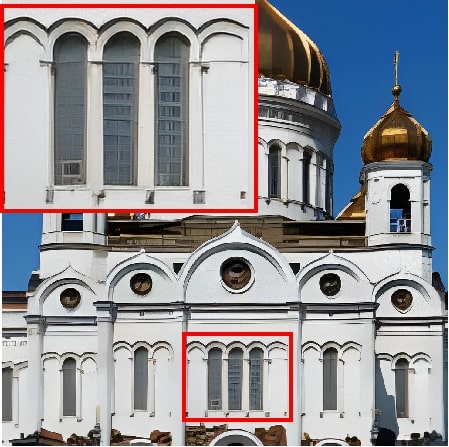}
      
    \end{subfigure}
    \begin{subfigure}[b]{0.16\hsize}
        \centering
        \includegraphics[width=\hsize]{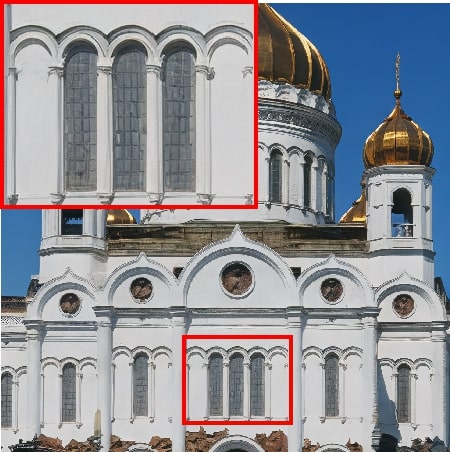}
 
    \end{subfigure}
    \begin{subfigure}[b]{0.16\hsize}
        \centering
        \includegraphics[width=\hsize]{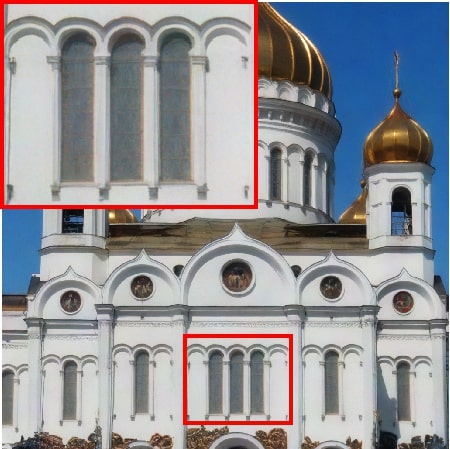}
 
    \end{subfigure}
    \begin{subfigure}[b]{0.16\hsize}
        \centering
        \includegraphics[width=\hsize]{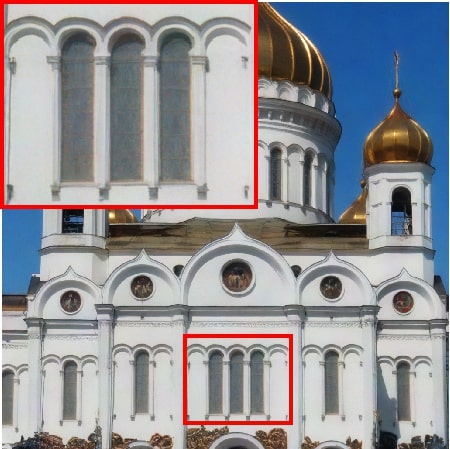}
 
    \end{subfigure}
    \begin{subfigure}[b]{0.16\hsize}
        \centering
        \includegraphics[width=\hsize]{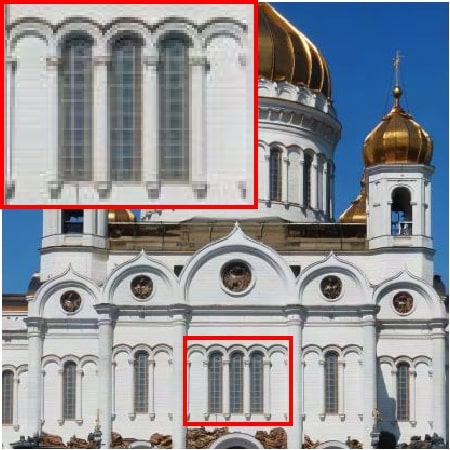}
 
    \end{subfigure}

        \vspace{-0.3cm} 
    \begin{subfigure}[b]{0.16\hsize}
        \centering
        \includegraphics[width=\hsize]{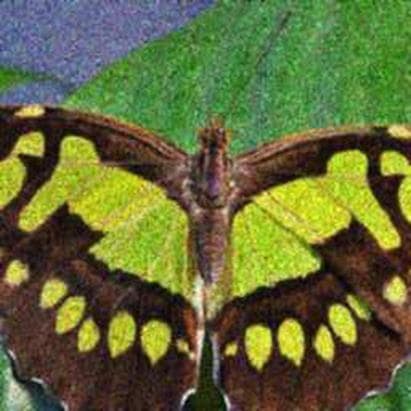}
   
    \end{subfigure}
    \begin{subfigure}[b]{0.16\hsize}
        \centering
        \includegraphics[width=\hsize]{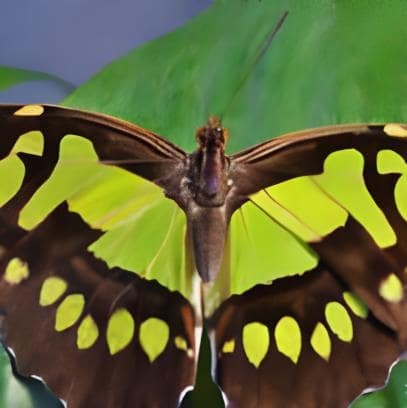}
      
    \end{subfigure}
    \begin{subfigure}[b]{0.16\hsize}
        \centering
        \includegraphics[width=\hsize]{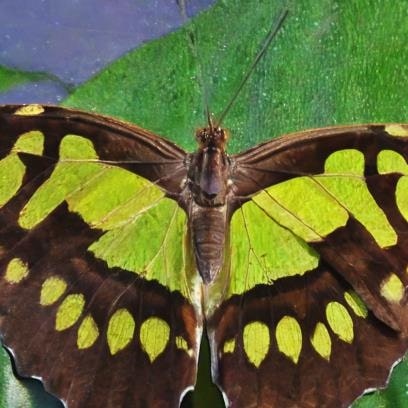}
 
    \end{subfigure}
    \begin{subfigure}[b]{0.16\hsize}
        \centering
        \includegraphics[width=\hsize]{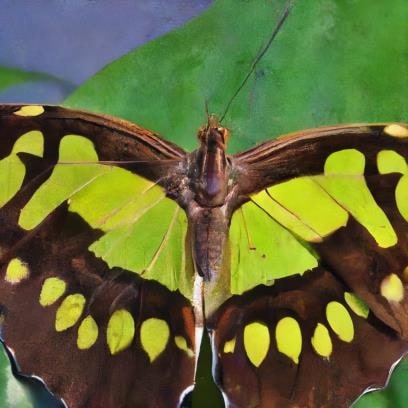}
 
    \end{subfigure}
    \begin{subfigure}[b]{0.16\hsize}
        \centering
        \includegraphics[width=\hsize]{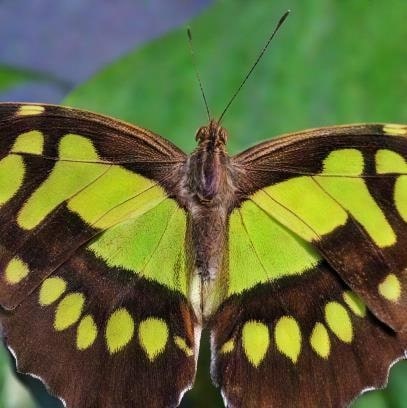}
 
    \end{subfigure}
    \begin{subfigure}[b]{0.16\hsize}
        \centering
        \includegraphics[width=\hsize]{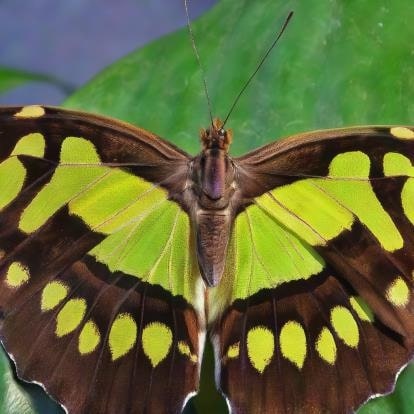}
 
    \end{subfigure}

        \vspace{-0.3cm} 
    \begin{subfigure}[b]{0.16\hsize}
        \centering
        \includegraphics[width=\hsize]{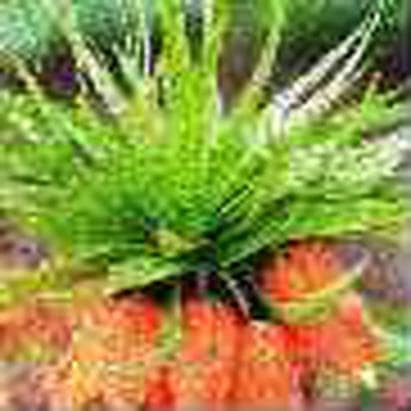}
        \caption*{LQ}
   
    \end{subfigure}
    \begin{subfigure}[b]{0.16\hsize}
        \centering
        \includegraphics[width=\hsize]{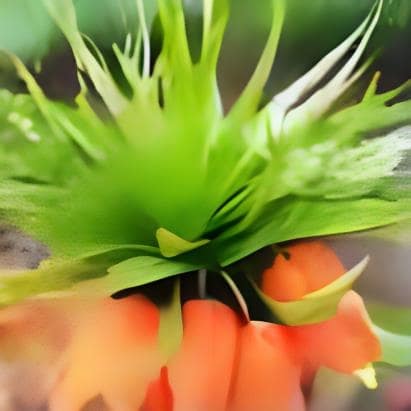}
        \caption*{Real-ESRGAN~\cite{wang2021real}}
      
    \end{subfigure}
    \begin{subfigure}[b]{0.16\hsize}
        \centering
        \includegraphics[width=\hsize]{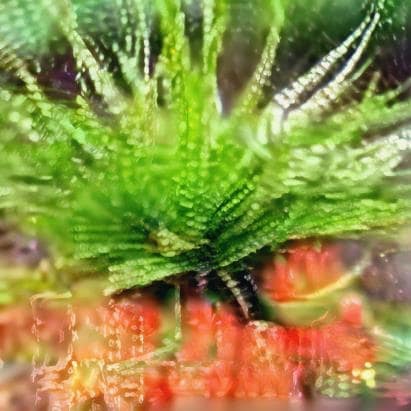}
        \caption*{StableSR~\cite{wang2024exploiting}}
 
    \end{subfigure}
    \begin{subfigure}[b]{0.16\hsize}
        \centering
        \includegraphics[width=\hsize]{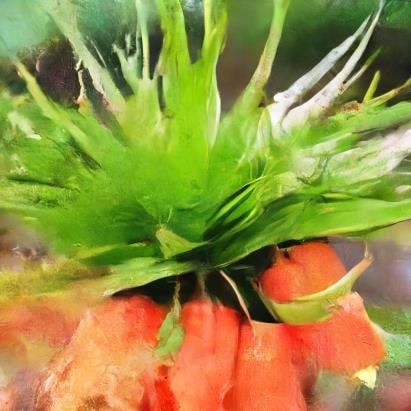}
        \caption*{SinSR~\cite{wang2024sinsr}}
 
    \end{subfigure}
    \begin{subfigure}[b]{0.16\hsize}
        \centering
        \includegraphics[width=\hsize]{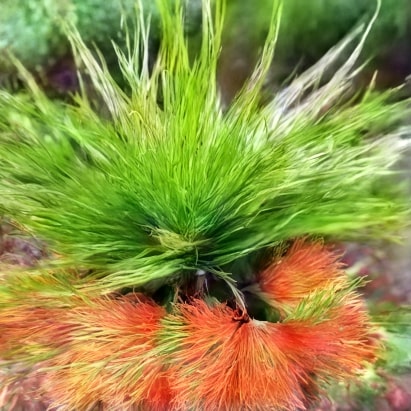}
        \caption*{SUPIR~\cite{yu2024scaling}}
 
    \end{subfigure}
    \begin{subfigure}[b]{0.16\hsize}
        \centering
        \includegraphics[width=\hsize]{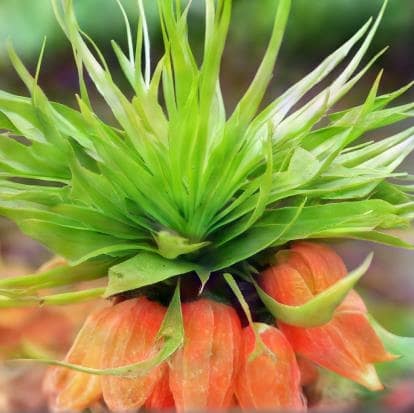}
        \caption*{DPIR(Ours)}
 
    \end{subfigure}

    \caption{Qualitative comparisons of different IR methods on DIV2K dataset. Our DPIR achieves the best visual performance.}
    \label{fig:compare1}

    \vspace{-4mm}
\end{figure*}

\subsection{Dual Prompting Control Branch}
\label{subsec:dpcb}
SD3 excels in high-quality image synthesis due to its strong multimodal language understanding, enabled by rich text embeddings and text-image cross-attention~\cite{esser2024scaling}.
In detial, SD3 includes three types of text encoders: two CLIP~\cite{radford2021learning} encoders and one T5~\cite{raffel2020exploring} encoder, with the final textual embeddings formed by concatenating the token embeddings of these three encoders.
Additionally, the pooled embedding from the two CLIP text encoders, which captures global semantics, is also incorporated into DiT.

 Inspired by the hybrid textual embeddinds of SD3, we design a dual prompting module that effectively expliots the local and global visual understanding perspective from the input low-quality image, along with the textual prompt input, as illustrated in Figure~\ref{fig:dual}.

\subsubsection{Dual Prompting Visual Control}
In image restoration, the low-quality image, $x_{\text{LQ}}$, serves as a crucial prior for restoring its high-fidelity image.
Visual details, such as structures and textures, are difficult to be accurately described using text. Moreover, DiT lacks the skip connections in U-Net, so the lightweight LQ image conditioning in Sec.~\ref{subsec:lqicb} alone is not sufficient to preserve LQ image features. To address this, we incorporate more visual control by replacing the CLIP text embeddings in DiT, which guides the restoration process more effectively.

Given a LQ image to be restored, specifically $x^{\text{local}}_{\text{LQ}}$ for a local image patch during training, we extract its visual features using CLIP image encoders as conditional control.
For simplicity, in the following, we omit the LQ subscript from $x^{\text{local}}_{\text{LQ}}$, referring to it as $x^{\text{local}}$.
In detail, we feed $x^{\text{local}}$ into two CLIP image encoders\footnote{https://huggingface.co/openai/clip-vit-large-patch14} \footnote{https://huggingface.co/openai/clip-vit-base-patch32}to obtain their visual class token embeddings as the pooled embeddings $c_{\text{pool}}$, and also into the first CLIP image encoder to extract its hidden states as the visual embeddings, $c^{\text{vis}}_{\text{local}}$.
These two visual prompt features capture the visual information of the LQ image patch from different receptive fields, replacing the original CLIP text embeddings in DiT.

\begin{table*}[tbp]
\centering
\small
\resizebox{\linewidth}{!}{
\begin{tabular}{@{}c|c|cccccccc@{}}
\toprule
Datasets                   & Methods             & PSNR↑ & SSIM↑  & LPIPS↓ & DISTS↓     &CLIPIQA↑   & MUSIQ↑ & NIQE↓ &MANIQA↑  \\ \midrule
\multirow{6}{*}{DRealsr}    &  BSRGAN  & 26.50&0.6917&	0.6334&	0.3603&	0.2402&	25.22&	7.7990&	0.2723   \\
                           & Real-ESRGAN         & \bd{28.05}&	\rf{0.7916}&	\bd{0.3875}	&\bd{0.2514}&	\underline{0.5593}&	\underline{58.84}&	\bd{5.8095}&	\rf{0.3883}
   \\
                           & StableSR            & \rf{28.12}&	\bd{0.7735}&	\rf{0.3634}	&\underline{0.2270}&0.4812&	48.81&	6.2847&	0.2782
   \\
                           & SinSR             & 26.83& 	0.6352& 	0.5849	& 0.2877	& \bd{0.6195}& 	56.57	& 7.4154	& \underline{0.3618}
   \\
                           & SUPIR     &      25.22&	0.6143&	0.4992&	0.2613&	0.5552&	\bd{59.02}&	\underline{5.9328}&	0.3607
   \\
                           & Ours     &  \underline{27.31} 	&\underline{0.7090} & \underline{0.3903}&\rf{0.1998}&	\rf{0.6424}&	\rf{62.95}	&\rf{5.5463}	&\bd{0.3880}

   \\ \bottomrule
\multirow{6}{*}{Realsr}    &  BSRGAN &22.03&	0.3723&	0.7722&	0.3795&	0.3951	&43.58	&6.0722&0.3224
   \\
                           & Real-ESRGAN      &  \bd{26.06}	&\rf{0.7700}&	\underline{0.2827}&	\bd{0.1936}&	0.5157&	\bd{64.46}&	\rf{4.4546}	&\underline{0.3879}
   \\
                           & StableSR            & \rf{26.35}&	\bd{0.7407}	&\bd{0.3083}&\underline{0.1990}&	\underline{0.5557}	&60.54	&4.9550&	0.3415
  \\
                           & SinSR           &   25.62 &	0.6777 &	0.4036	 &0.2384 &\rf{0.6678}	 &\underline{64.25} &	6.0367 &	\bd{0.4391}

  \\
                           & SUPIR           & 25.20&	0.6423&	0.3996&	0.2268	&0.5223	&58.68	&\underline{5.3074}	&0.3570   \\
                           & Ours          & \bd{26.06}&	\underline{0.7371}&	\rf{0.2641}	&\rf{0.1642}&	\bd{0.6625}&	\rf{69.28}&	\bd{4.7744}	&\rf{0.4857}

   \\ \bottomrule
\multirow{6}{*}{DIV2K-Val}    &  BSRGAN & \underline{22.40}&	0.5352&	0.7052&	0.3597&0.2228&	23.24	&7.2941&	0.2224
   \\
                           & Real-ESRGAN         & \bd{22.62}&	\rf{0.6025}&	\underline{0.3982}	&0.2240	&\underline{0.5661}&	63.90&	\bd{3.9423}&	\underline{0.3782}
   \\
                           & StableSR          &   \rf{22.87}	& \bd{0.5844}	& \bd{0.3925}	& \underline{0.2085}& 	0.4974	& 57.28	& 4.5447	& 0.2848
  \\
                           & SinSR            & 22.10	&0.5212&	0.4416&	0.2160&	\bd{0.6919}&	\underline{65.13}	&5.7650&	\bd{0.4178}
   \\
                           & SUPIR      &     21.23	&0.5075&0.4152&\bd{0.1873}&	0.5239&	\bd{66.49}&	\rf{3.7168}&	0.3370
   \\
                           & Ours       &   21.61& 	\underline{0.5379}	& \rf{0.3622}	& \rf{0.1677}	& \rf{0.7416}& 	\rf{71.94}	& \underline{3.9847}	& \rf{0.5330}

   \\ \bottomrule
\multirow{6}{*}{CelebA}    &  BSRGAN& 22.97	&0.3347&	0.7168	&0.3311	&0.5384	&44.43&	5.7392&0.3574
   \\
                           & Real-ESRGAN       &  \rf{30.34}	&\rf{0.8211}&	\underline{0.3135}&	\underline{0.1743}&	0.5043&	63.94&	5.4870	&0.3285
   \\
                           & StableSR            & \bd{29.61}& \bd{0.7932}& 	\bd{0.2925}& 	\bd{0.1618}& 	0.6394& 	66.88& 	\underline{5.1979}& 	0.3504
  \\
                           & SinSR             & 29.19	 &0.7465 &	0.3469 &	0.1807 &\rf{0.7940} &	\underline{71.43}	 &6.2762 &	\bd{0.4348}
   \\
                           & SUPIR           & 27.17& 	0.6479& 	0.3929	& 0.1943	& \underline{0.7014}& 	\rf{74.57}	& \bd{4.9066}	& \underline{0.4188}
  \\
                           & Ours          & \underline{29.58}&	\underline{0.7739}	&\rf{0.2449}	&\rf{0.1302}	&\bd{0.7111}&	\bd{71.49}	&\rf{4.8288}	&\rf{0.4414}
   \\ \bottomrule

\end{tabular}}
\caption{Quantitative comparison with state-of-the-art methods on synthetic datasets. The best, second-best, and third-best results for each metric are highlighted in
\textcolor{red}{\textbf{red}}, \textcolor{blue}{\textbf{blue}}, and \underline{underlined}, respectively.}
\label{tab:main}
\vspace{-4mm}
\end{table*}

\subsubsection{Global-Local Visual Training}
Image restoration often requires high-resolution outputs, such as 2K or 4K.
Local image patches in such large images, such as only the local view of a dog's eye, may not reveal the object or overall context of the image to be restored.
This contrasts with the text control signals in pre-trained T2I models, where text captures more global semantics, which potentially leads to learning difficulties during training and reduced final performance.
To address this gap, we propose a global-local visual understanding training strategy to smoothly adapt pre-trained DiT from image synthesis to image restoration.

As shown in Figure~\ref{fig:dual}, during training, given a high-resolution image, we crop additional global context patches, $x^{\text{global}}$, surrounding the local patch $x^{\text{local}}$ to be restored.
These global patches are also fed into the first CLIP image encoder to produce the global visual features $c^{\text{vis}}_{\text{global}}$.
These global visual tokens capture more contextual and semantic information from the input image, making them more similar to the representations of the original text embeddings in SD3.
Then, we concatenate $c^{\text{vis}}_{\text{global}}$ with previously obtained local visual tokens $c^{\text{vis}}_{\text{local}}$ to obtan the final global-local visual prompts $c^{\text{vis}}_{\text{G-L}}$.
Next, the visual prompts, combined with the text prompts from T5, result in the dual prompt $c_{\text{dual}}$, which is applied to each block in DiT via cross-attention.
To enhance the model's generalization ability across different resolutions, DPIR is trained on images with a combination of resolutions (\eg, 1k, 2k, 4k).
Therefore, DPIR can set $x^{\text{global}}$ to $x^{\text{local}}$ when the $x^{\text{global}}$ is not large enough to be split into patches for low-resolution inference.

\begin{figure*}[t]
    \centering

        \vspace{-0.3cm} 
    \begin{subfigure}[b]{0.16\hsize}
        \centering
        \includegraphics[width=\hsize]{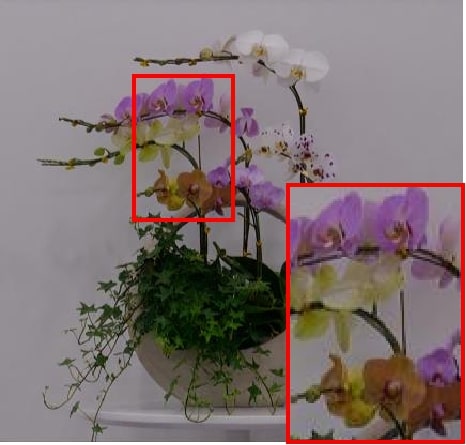}
      
    \end{subfigure}
    \begin{subfigure}[b]{0.16\hsize}
        \centering
        \includegraphics[width=\hsize]{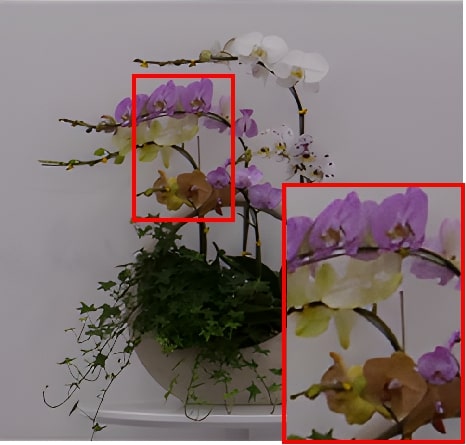}
      
    \end{subfigure}
    \begin{subfigure}[b]{0.16\hsize}
        \centering
        \includegraphics[width=\hsize]{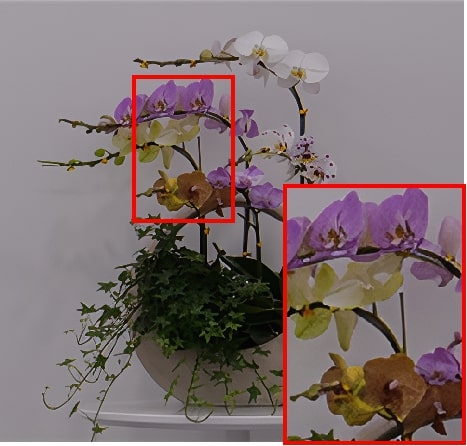}
 
    \end{subfigure}
    \begin{subfigure}[b]{0.16\hsize}
        \centering
        \includegraphics[width=\hsize]{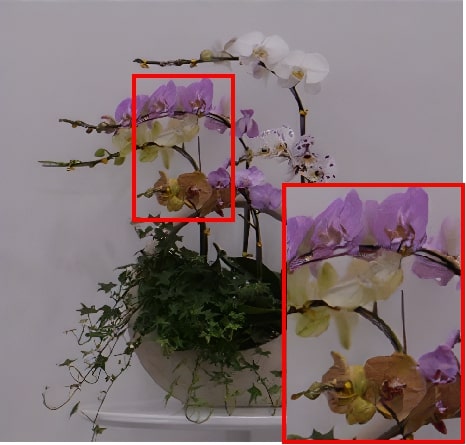}
 
    \end{subfigure}
    \begin{subfigure}[b]{0.16\hsize}
        \centering
        \includegraphics[width=\hsize]{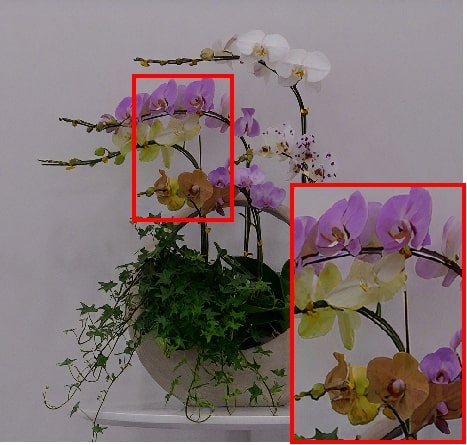}
 
    \end{subfigure}
    \begin{subfigure}[b]{0.16\hsize}
        \centering
        \includegraphics[width=\hsize]{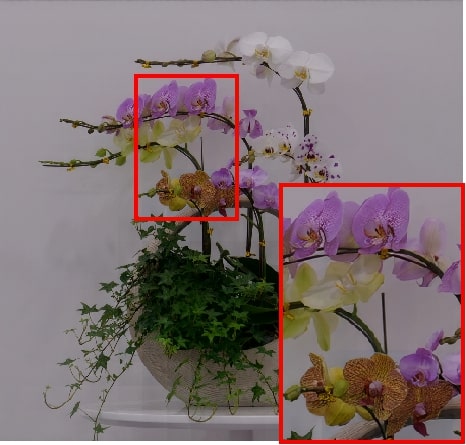}
 
    \end{subfigure}

        \vspace{-0.3cm} 
    \begin{subfigure}[b]{0.16\hsize}
        \centering
        \includegraphics[width=\hsize]{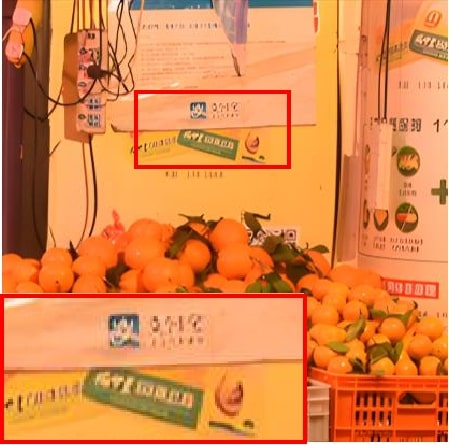}
        \caption*{LQ}
      
    \end{subfigure}
    \begin{subfigure}[b]{0.16\hsize}
        \centering
        \includegraphics[width=\hsize]{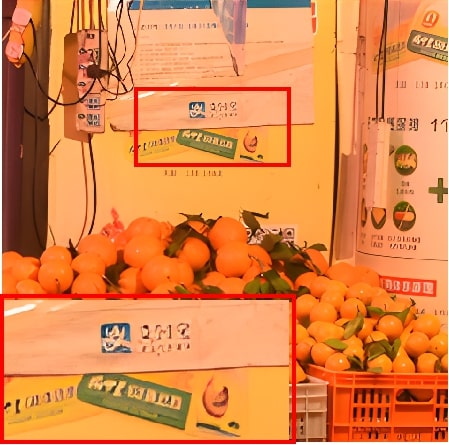}
        \caption*{Real-ESRGAN~\cite{wang2021real}}
      
    \end{subfigure}
    \begin{subfigure}[b]{0.16\hsize}
        \centering
        \includegraphics[width=\hsize]{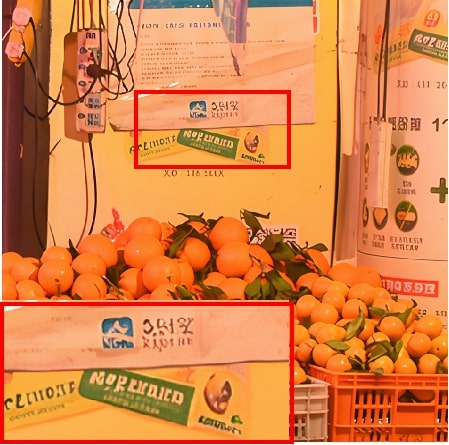}
        \caption*{StableSR~\cite{wang2024exploiting}}
 
    \end{subfigure}
    \begin{subfigure}[b]{0.16\hsize}
        \centering
        \includegraphics[width=\hsize]{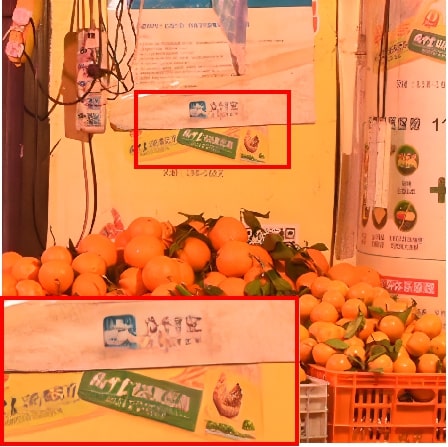}
        \caption*{SinSR~\cite{wang2024sinsr}}
 
    \end{subfigure}
    \begin{subfigure}[b]{0.16\hsize}
        \centering
        \includegraphics[width=\hsize]{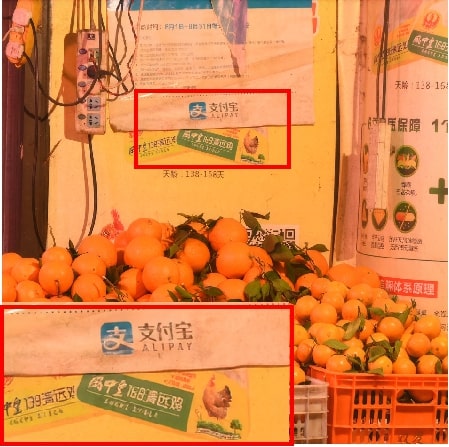}
        \caption*{SUPIR~\cite{yu2024scaling}}
 
    \end{subfigure}
    \begin{subfigure}[b]{0.16\hsize}
        \centering
        \includegraphics[width=\hsize]{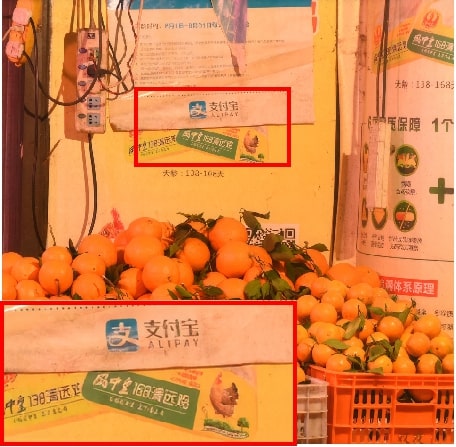}
        \caption*{DPIR (Ours)}
 
    \end{subfigure}

    \caption{Qualitative comparisons of real images. DPIR delivers outstanding restoration performance, particularly in preserving details.}
    \label{fig:real}
    \vspace{-3mm}
\end{figure*}

\section{Experiments}
\subsection{Experimental Settings}
\textbf{Datasets.}
To meet the requirements of the base DiT, we crop images into 1024×1024 patches during training.
LQ images are generated by the synthetic degradation method, following Real-ESRGAN~\cite{wang2021real}.
For evaluation on synthetic data, we evaluate DPIR on DIV2K-val~\cite{agustsson2017ntire}, DRealSR~\cite{wei2020component}, RealSR~\cite{cai2019toward} and CelebA~\cite{liu2015deep}.
We perform degradation on them following~\cite{yu2024scaling,wang2021real}.
All datasets are cropped to 1024×1024 and degraded to 256×256 LR images.
For evaluation on real-world data, we choose DRealSR and crop the HQ images to 1024x1024 for comparison purposes. 

\noindent \textbf{Implementation Details. }
For training degradation-robust VAE, we use the AdamW optimizer with a learning rate of $1\times10^{-5}$. We employ the same training loss as VQGAN~\cite{Esser2020TamingTF}, but only fine-tune the VAE encoder.
During the first 5000 training steps, we use L1 loss and LPIPS~\cite{zhang2018unreasonable} loss, and then we use a discriminator and train for additional 20,000 steps.
To train the DPIR model, including the lightweight module of the image conditioning branch, the DiT backbone, and the MLP layers of the dual prompting module, we first pre-train it on a self-collected dataset using a batch size of 1024 and a learning rate of $1\times10^{-4}$.
Next, we fine-tune it with a batch size of 256 and a learning rate of $2\times10^{-5}$.
This two-stage training ensures that the model benefits from a large, diverse dataset initially and further improves its detail generation performance with higher-quality data.

\subsection{Comparison and Evaluation}
We compare our method with several state-of-the-art methods.
GAN-based methods include BSRGAN~\cite{zhang2021designing}, Real-ESRGAN~\cite{wang2021real}.
Diffusion-based methods include StableSR~\cite{wang2024exploiting}, SinSR~\cite{wang2024sinsr}, and SUPIR~\cite{yu2024scaling}.
We adopt commonly-used no-reference metrics (\ie, CLIP-IQA~\cite{wang2023exploring}, MUSIQ~\cite{ke2021musiq}, NIQE~\cite{mittal2012making}, MANIQA~\cite{yang2022maniqa}) and reference metrics (\ie, PSNR, SSIM, LPIPS, DISTS).


\vspace{-2mm}
\begin{table}[tbp]
\setlength{\abovecaptionskip}{0.1cm}
\setlength{\belowcaptionskip}{-0.1cm}
\centering
\footnotesize
\resizebox{\linewidth}{!}{
\begin{tabular}{@{}c|cccc@{}}
\toprule
              & LPIPS↓ &DISTS↓  &CLIPIQA↑ & MUSIQ↑ \\ \midrule
DACLIP-UIR & 0.5686 & 0.3528 &0.3526   &	47.50  \\
DPIR (ours)  &\rf{0.3903}&\rf{0.1998}&\rf{0.6424} &	\rf{62.95}
   \\ \bottomrule
\end{tabular}}
\caption{Comparison with DACLIP on DrealSR.}
\label{tab:daclip}
\end{table}

\begin{table}[htbp]
\setlength{\abovecaptionskip}{0.1cm}
\setlength{\belowcaptionskip}{-0.1cm}
\centering
\footnotesize
\resizebox{\linewidth}{!}{
\begin{tabular}{@{}c|cccccccc@{}}
\toprule
         & PSNR↑ & LPIPS↓ &CLIPIQA↑   & MUSIQ↑  \\ \midrule
DACLIP&32.84 &	0.1983	&0.4363
	&	46.33
 \\
 DPIR (ours)&\rf{35.46}&\rf{0.1808}&\rf{0.5150}&\rf{57.96}

   \\ \bottomrule
\end{tabular}}
\caption{Comparison with DACLIP on PolyU.}
\label{tab:reb_denoising}
\end{table}

\begin{table}[t]
\setlength{\abovecaptionskip}{0.1cm}
\setlength{\belowcaptionskip}{-0.1cm}
\centering
\footnotesize
\resizebox{\linewidth}{!}{
\begin{tabular}{@{}c|cccc@{}}
\toprule
 Methods               &CLIPIQA↑   & MUSIQ↑ & NIQE↓ &MANIQA↑ \\ \midrule
 BSRGAN & 0.5994&	62.19	&\bd{3.9264}&	0.3555
   \\
  Real-ESRGAN      &   0.5749&	67.10&4.3671	&0.4155
  \\
 StableSR         &   0.6594	&\bd{68.58}	&5.0153&	0.4327
   \\
                           
 SinSR       & \bd{0.6842}	&67.92	&4.9608&	\bd{0.4730}
   \\
 SUPIR           & 0.5632&63.93&5.2670&	0.4150
   \\ 
   Ours & \rf{0.6980}	&\rf{70.85}&	\rf{3.7382}&	\rf{0.5310}
\\
   \bottomrule
\end{tabular}
}
\caption{Quantitative comparison with state-of-the-art methods on real-world benchmarks.}
\label{real}
\end{table}

\begin{figure}[htbp]
\setlength{\abovecaptionskip}{0.2cm}
\setlength{\belowcaptionskip}{-0.15cm}
    \centering
    \begin{subfigure}[b]{0.315\hsize}
        \centering
        \includegraphics[width=\hsize]{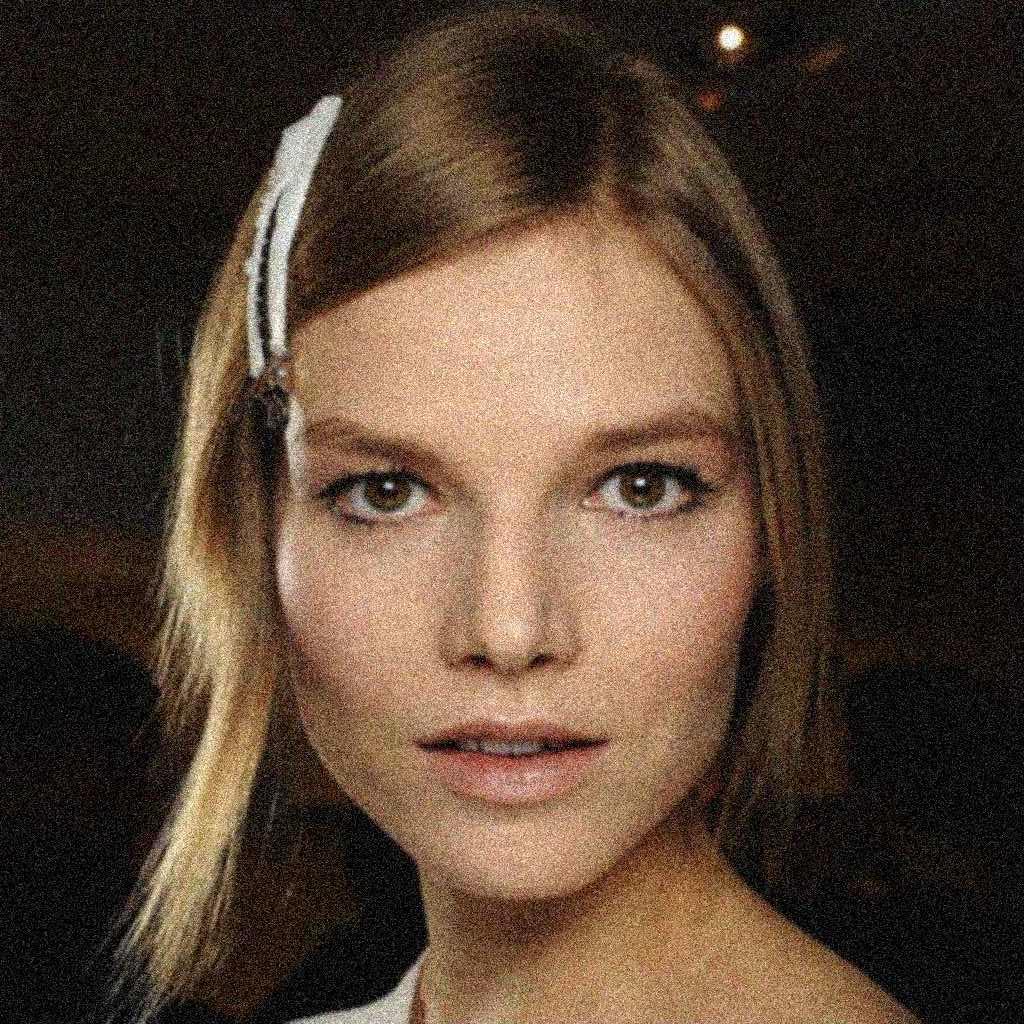}
        \caption*{LQ}
    \end{subfigure}
    \begin{subfigure}[b]{0.315\hsize}
        \centering
        \includegraphics[width=\hsize]{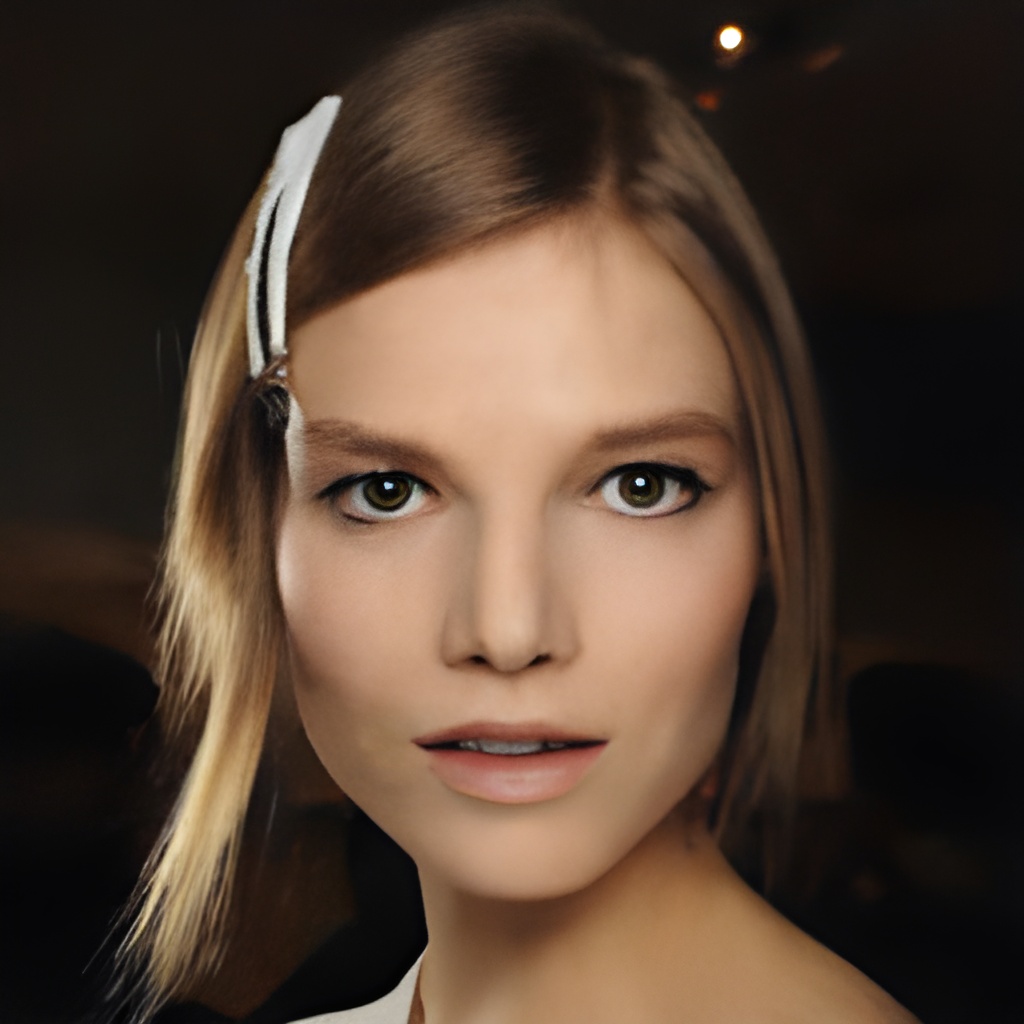}
        \caption*{Real-ESRGAN~\cite{wang2021real}}
    \end{subfigure}
    \begin{subfigure}[b]{0.315\hsize}
        \centering
        \includegraphics[width=\hsize]{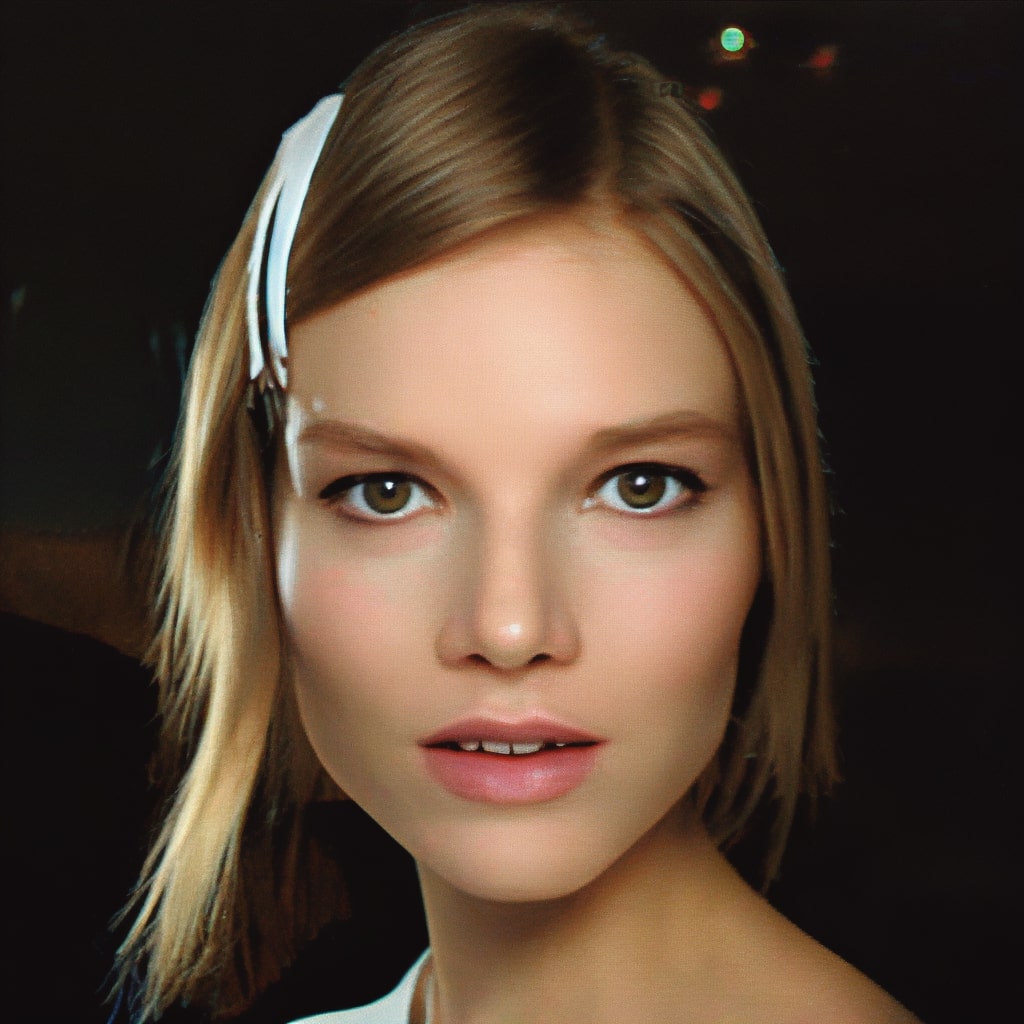}
        \caption*{StableSR~\cite{wang2024exploiting}}
    \end{subfigure}
    
    \vspace{0.1cm} 
    \begin{subfigure}[b]{0.315\hsize}
        \centering
        \includegraphics[width=\hsize]{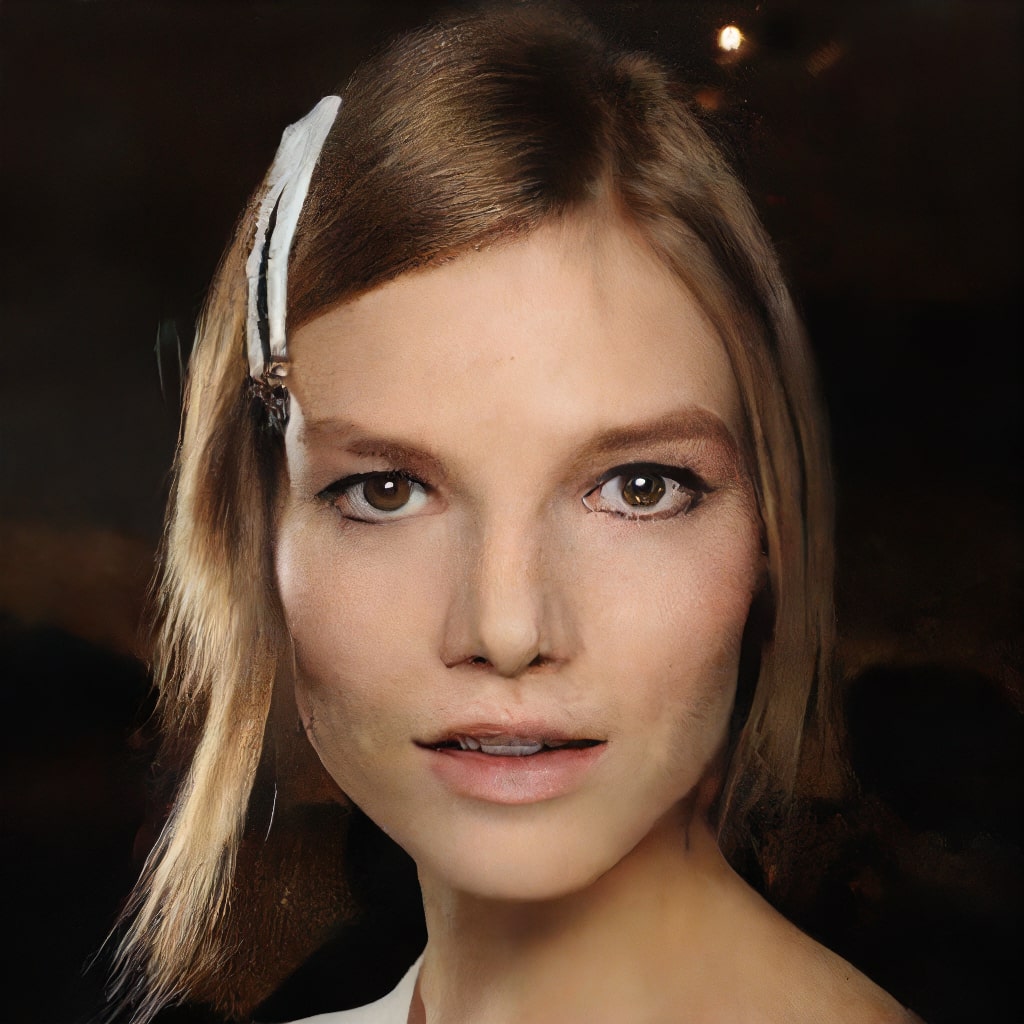}
        \caption*{SinSR~\cite{wang2024sinsr}}
        \label{fig:sub4}
    \end{subfigure}
    \begin{subfigure}[b]{0.315\hsize}
        \centering
        \includegraphics[width=\hsize]{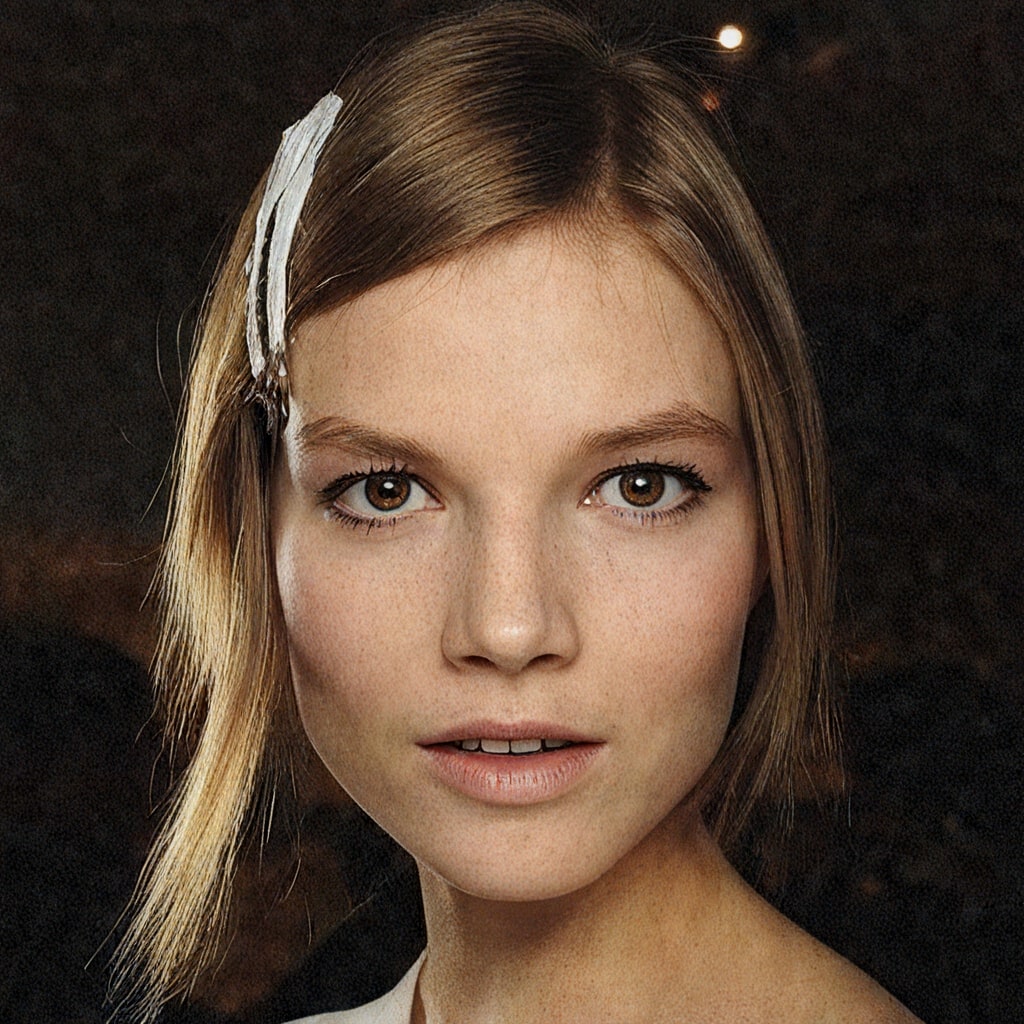}
        \caption*{SUPIR~\cite{yu2024scaling}}
        \label{fig:sub5}
    \end{subfigure}
    \begin{subfigure}[b]{0.315\hsize}
        \centering
        \includegraphics[width=\hsize]{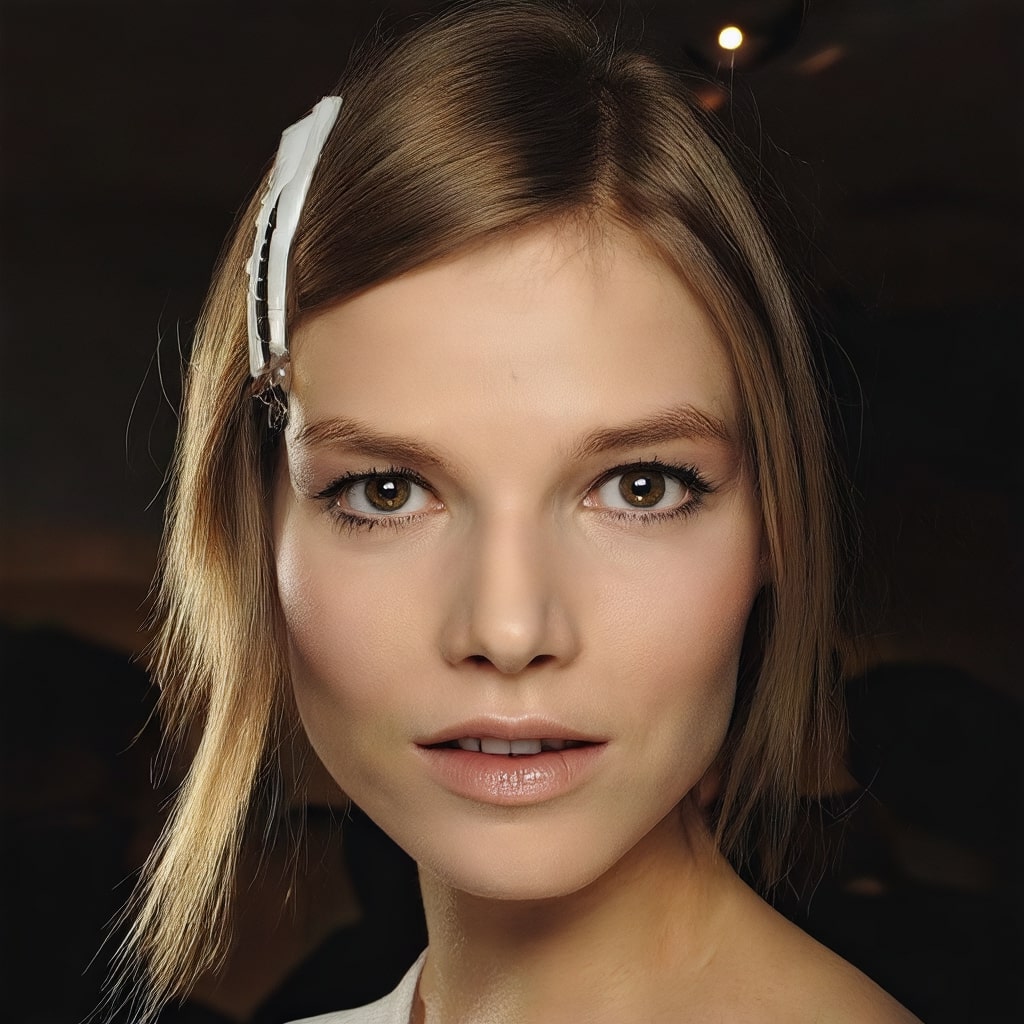}
        \caption*{DPIR (Ours)}
        \label{fig:sub6}
    \end{subfigure}
    \caption{Qualitative comparisons on face images.}
    \vspace{-4mm}
    \label{fig:face}
\end{figure}

\begin{figure*}[tp]
\setlength{\abovecaptionskip}{0.4cm}
\setlength{\belowcaptionskip}{-0.1cm}
    \centering
    \begin{subfigure}[b]{0.145\hsize}
        \centering
        \includegraphics[width=\hsize]{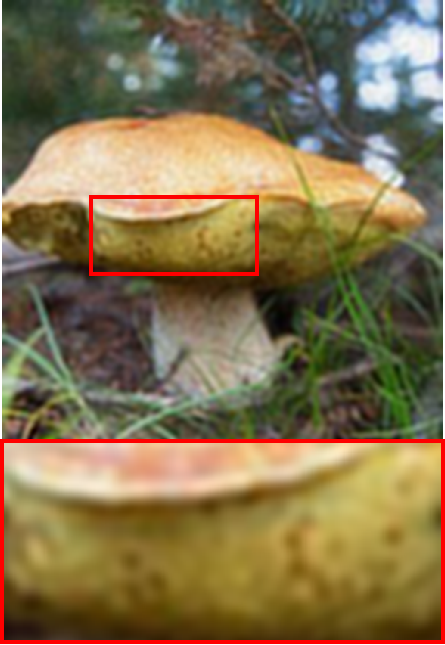}
        \caption*{LQ}
      
    \end{subfigure}
    \begin{subfigure}[b]{0.145\hsize}
        \centering
        \includegraphics[width=\hsize]{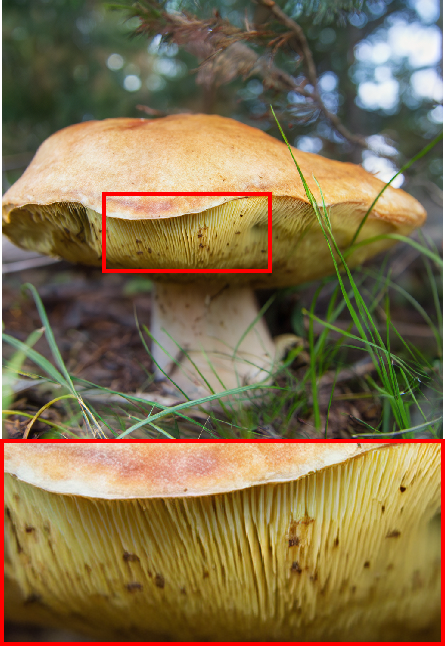}
        \caption*{Textual Prompts}
      
    \end{subfigure}
    \begin{subfigure}[b]{0.145\hsize}
        \centering
        \includegraphics[width=\hsize]{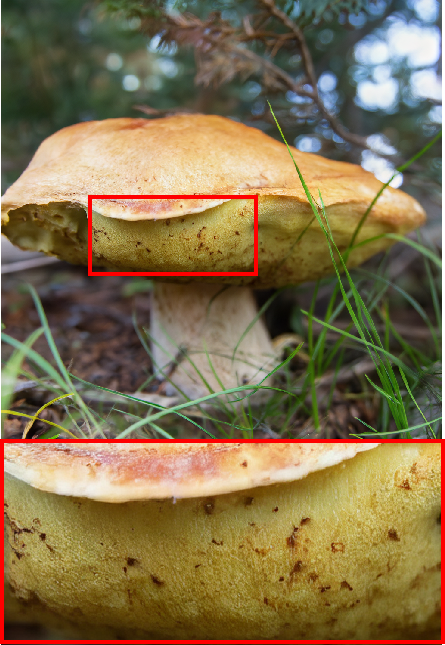}
        \caption*{Dual Prompts}
  
    \end{subfigure}
    \begin{subfigure}[b]{0.18\hsize}
        \centering
        \includegraphics[width=\hsize]{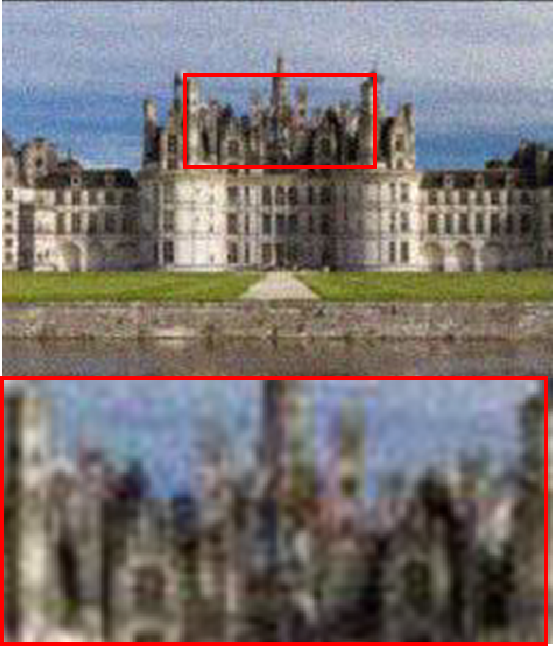}
        \caption*{LQ}
  
    \end{subfigure}
    \begin{subfigure}[b]{0.18\hsize}
        \centering
        \includegraphics[width=\hsize]{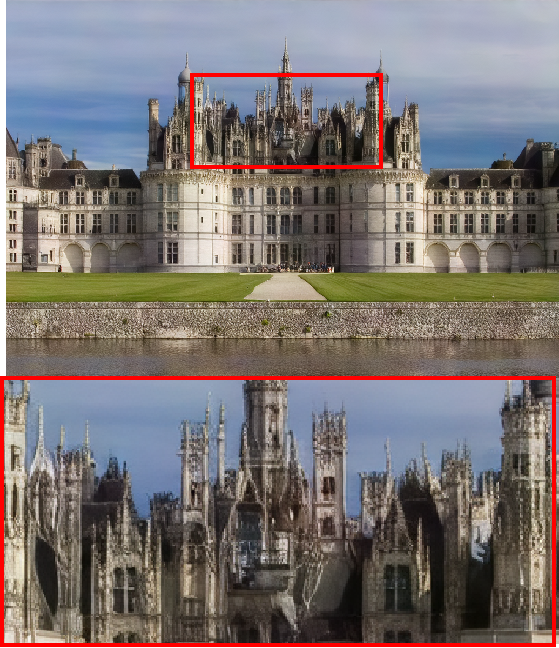}
        \caption*{Textual Prompts}
  
    \end{subfigure}
    \begin{subfigure}[b]{0.18\hsize}
        \centering
        \includegraphics[width=\hsize]{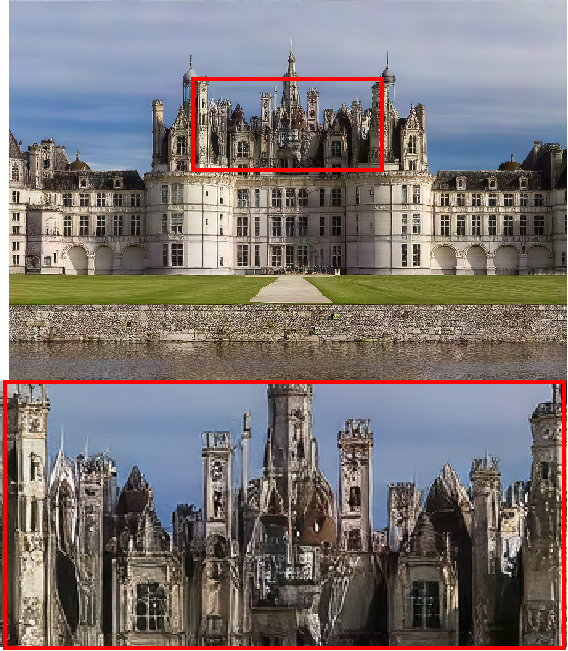}
        \caption*{Dual Prompts}
  
    \end{subfigure}

    \caption{Qualitative comparisons of different control prompting strategies. Our dual prompting achieves the best visual results.}
    \label{fig:dual3}
\end{figure*}

\begin{table*}[t]
\centering
\setlength{\abovecaptionskip}{0.1cm}
\setlength{\belowcaptionskip}{-0.5cm}
\footnotesize
\begin{tabular}{@{}c|cccccccc@{}}
\toprule
 Methods             & PSNR↑ & SSIM↑  & LPIPS↓ & DISTS↓     &CLIPIQA↑   & MUSIQ↑ & NIQE↓ &MANIQA↑  \\ \midrule
Text Prompts Only & 27.21&0.7047&0.4217&0.2173&0.5614&57.08&5.7048&0.3446
 \\
Visual Prompts Only  & \rf{28.69}&\rf{0.7239}&0.6239&0.3739&0.4106&40.97&12.6444&0.29278
   \\
Dual Prompts  &27.63
&0.7164&\rf{0.3846}&\rf{0.1981}&\rf{0.6175}&\rf{60.30}&\rf{5.5820
}&\rf{0.3650}

   \\ \bottomrule

\end{tabular}
\caption{Quantitative comparison with Text Prompting, Image Prompting and Dual Prompting.}
\label{tab:ablation_dual1}
\end{table*}

\subsubsection{Evaluation on Synthetic Data}
Table \ref{tab:main} shows the quantitative comparison results on three datasets of synthetic data.
Our method consistently achieves the best or the second-best scores of DISTS, and no-reference metrics across all three datasets. 
In Realsr and DIV2K dataset, our method has the best score of LPIPS.
In particular, our method achieves a significant lead in no-reference metrics. 

Figure \ref{fig:compare1} shows qualitative comparison results, illustrating the superiority of our method in terms of visual effects.
Even under severe degradation, our method still maintains good fidelity and shows strong detail reconstruction capabilities.
Most diffusion model-based methods struggle to maintain semantic consistency with the low-quality image, often producing incorrect textures.
Our method accurately restores details in various scenes, such as the structures of animals, windows of buildings, and leaves of plants.

We also compared DPIR and DACLIP in SR and denoising tasks. Table \ref{tab:daclip} and Table \ref{tab:reb_denoising} show our strong performance. Furthermore, to demonstrate that our method performs well on specific image categories, we validate its performance on a face dataset without any additional fine-tuning.
The quantitative metrics and visual results are are presented in Table \ref{tab:main} and Figure \ref{fig:face}, respectively.
Our method achieves excellent results in facial restoration, outperforming other approaches in both quantitative metrics and visual quality.







\vspace{-1mm}
\subsubsection{Evaluation on Real-World Data}

We adopt DRealSR~\cite{wei2020component} dataset as the real-world LQ testset. 
The qualitative results are shown in Table \ref{real}, and the qualitive results are shown in Figure \ref{fig:real}.
In addition to achieving the best performance across all no-reference metrics, our visual results also surpass previous methods in terms of detail and fidelity.
These results demonstrate that our method performs exceptionally well on real data, effectively addressing complex real-world degradations across a wide range of scenarios.
We provide more visual comparisons and a user study in the appendix.

\begin{table}[t]
\centering
\setlength{\abovecaptionskip}{0.2cm}
\setlength{\belowcaptionskip}{-0.5cm}
\tiny
\resizebox{\linewidth}{!}{
\begin{tabular}{@{}c|cccc@{}}
\toprule
 Methods        &LPIPS       &CLIPIQA↑   & MUSIQ↑ &MANIQA↑ 
 \\ \midrule
 Local & 0.3622&	0.6730
	&69.37
&	0.4315

   \\
Global-Local     &   \rf{0.3284}

&	\rf{0.7416}
&\rf{71.94}
	&\rf{0.5330}

  \\

   \bottomrule
\end{tabular}
}
\caption{Quantitative comparison of global-local and local visual training strategies.}
\label{tab:global}
\vspace{-4mm}
\end{table}

\begin{figure}[t]
 \setlength{\abovecaptionskip}{0.1cm}
\setlength{\belowcaptionskip}{-0.01cm}
    \centering
    \includegraphics[width=\linewidth]{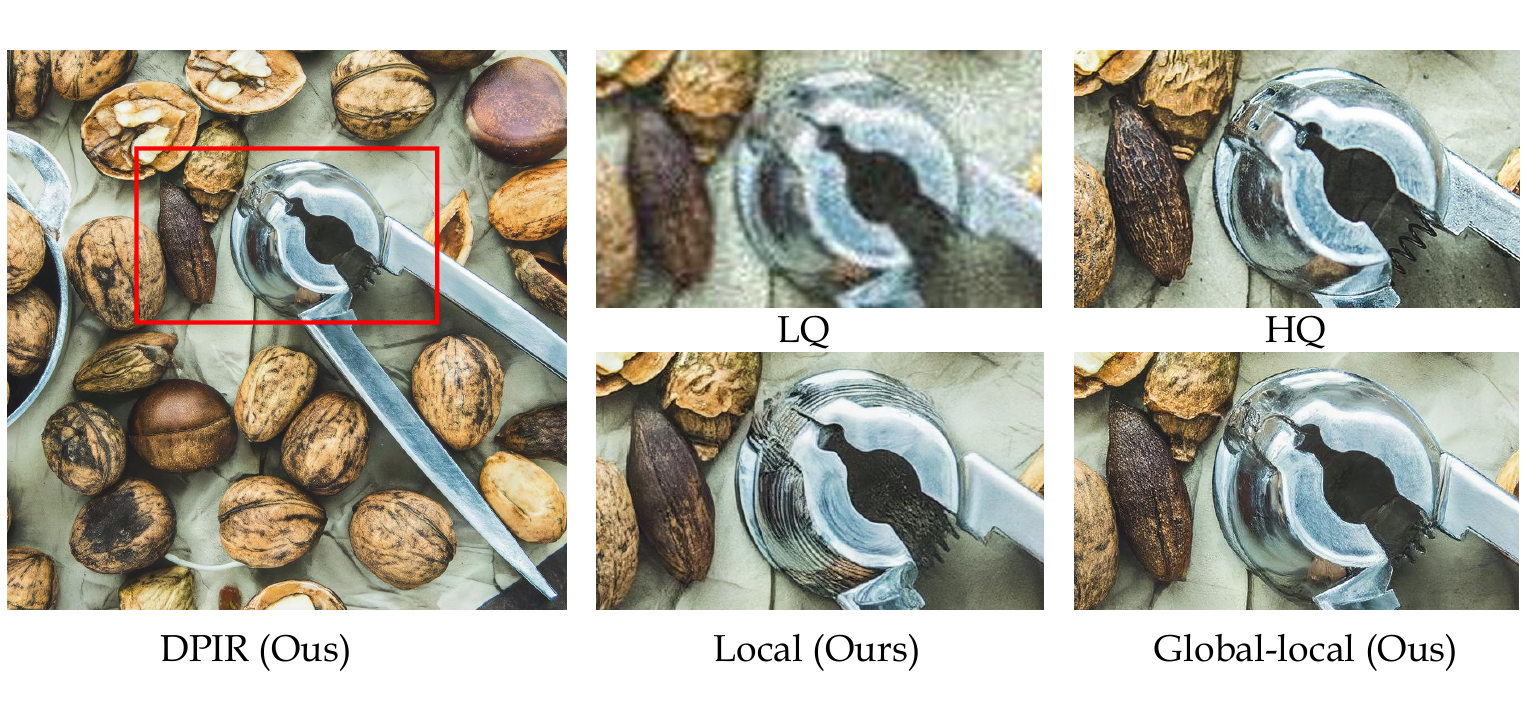}
    \vspace{-6mm}
     
    \caption{
    Visual comparisons of global-local and local methods.
    }
  
    \vspace{-6mm}
    \label{fig:global}
\end{figure}


\subsection{Ablation Study}
\label{sec:ablation}
\textbf{Prompting Strategy.}
We first discuss the significance of dual prompting in improving model performance. Specifically, we compare our dual prompting method against methods that rely solely on textual prompting or visual prompting. Textual prompting refers to the original SD3 implementation and visual prompting involves using only the extracted visual prompts from CLIP while removing T5 text prompts. As shown in Table \ref{tab:ablation_dual1}, dual prompting leads to improvements in nearly all metric scores, significantly enhancing the image restoration quality, particularly in terms of image aesthetics.
Figure \ref{fig:dual3} shows the visual effectiveness of the dual prompting method across different scenarios.
Note that the visual prompting variant lags largely behind text prompting (Table \ref{tab:ablation_dual1}), so we primarily compare against text prompting, the default setting in SD3.
We attribute this performance degradation to the absence of control signals, where text provides essential overall information about the scene or object to be restored.
Overall, our dual prompting, combining visual and textual information from the low-quality image, significantly enhances restoration outcomes.

\vspace{-4mm}
\paragraph{Global-Local Visual Training.}
We conduct ablation studies to evaluate the effectiveness of the global-local visual training strategy by performing experiments with and without extracting global-local visual features in dual prompting during training.
The ``Local'' variant extracts visual features solely from the local input patch, while ``Global-Local'' also leverages the global visual context from neighboring patches.
Table \ref{tab:global} and Figure \ref{fig:global} present the quantitative and qualitative results, respectively.
Our global-local visual training method greatly improve the restoration performance in terms of the image quality, substantially boosting scores across all metrics.
From the visual results, the model using this training method can faithfully restore details.
In contrast, the model trained using only local information tends to produce less accurate details.
We attribute this to a misunderstanding of the scene to be restored, as training solely with local visual information tends to focus on generating textures rather than preserving correct overall semantics.
Instead, during training, global-local visual information helps the model reduce the learning difficulty of restoring specific regions by providing context awareness.

\vspace{-2mm}
\section{Conclusion}
\label{sec:conclusion}
This paper presents DPIR, a novel DiT-based image restoration method and a new dual prompting strategy that extracts valuable visual conditioning from both global and local perspectives, significantly enhancing restoration outcomes. Experimental results demonstrate DPIR’s superior performance.

\paragraph{Acknowledgment.}

This work has been supported in part by National Natural Science
Foundation of China (Nos. 62322216, 62172409, 62311530686,U24B20175), research on the Optimization of Government Affairs Services (No. PBD2024-0521), research on the Detection and Analysis of Fake Threat Intelligence (No. C22600-15)

\cleardoublepage
{
    \small
    \bibliographystyle{ieeenat_fullname}
    \bibliography{main}
}


\end{document}